\def\includeallfigs{}

\documentclass[final,5p,times,sort&compress]{elsarticle}

\usepackage{amsmath,amssymb} %
\usepackage{color}

\graphicspath{{./figures/}{./figures/nestseq/}}

\usepackage{tabularx}
\newcolumntype{C}{>{\centering\arraybackslash}X}

\usepackage{verbatim}
\usepackage{url}
\usepackage[colorlinks=true]{hyperref}
\usepackage{stmaryrd}
\usepackage{multirow}
\usepackage{booktabs}
\usepackage{algorithm}
\usepackage{algorithmic}
\usepackage{subfigure}

\journal{Computer Vision and Image Understanding}

\begin{document}

\newcommand{\liff}{\leftrightarrow}

\newcommand{\proj}{\textup{proj}}

\newcommand{\R}{\mathbb{R}}
\newcommand{\N}{\mathbb{N}}
\newcommand{\Z}{\mathbb{Z}}
\newcommand{\Grid}{\mathcal{R}}
\newcommand{\Clique}{{\boldsymbol{\alpha}}}
\newcommand{\Freq}[1]{\mathbf{F}_{#1}}
\newcommand{\Hist}[1]{\vec{h}_{#1}}
\newcommand{\Image}{\mathbf{g}}
\newcommand{\Imagepix}{x}
\newcommand{\ImageSample}{\Image_\textup{samp}}
\newcommand{\ImageObs}{\Image^{\circ}}
\newcommand{\PiecesObs}{\{\ImageObs_i\}}
\newcommand{\barImageObs}{\bar{\Image}^{\circ}}
\newcommand{\V}{\mathbf{V}}
\newcommand{\NCC}{\textup{NCC}}
\newcommand{\TSS}{\textup{TSS}}
\newcommand{\JS}{\textup{JS}}
\newcommand{\KL}{\textup{KL}}

\newcommand{\params}{{\boldsymbol{\mathbf{\theta}}}}

\renewcommand{\vec}[1]{\mathbf{#1}}

\begin{frontmatter}

\title{Texture Modelling with Nested High--order Markov--Gibbs Random Fields \tnoteref{CCnote}}

\tnotetext[CCnote]{
  \copyright 2015. Licensed under Creative Commons CC-BY-NC-ND 4.0 \\ \url{http://creativecommons.org/licenses/by-nc-nd/4.0/}
}

\author{Ralph Versteegen}
\ead{rver017@aucklanduni.ac.nz}
\author{Georgy Gimel'farb}
\ead{g.gimelfarb@auckland.ac.nz}
\author{Patricia Riddle}
\ead{p.riddle@auckland.ac.nz}

\address{
  Department of Computer Science,
  The University of Auckland,\\
  Auckland 1142, New Zealand\\
}

\begin{abstract}
Currently, Markov--Gibbs random field (MGRF) image models
which include high-order interactions
are almost always
built by modelling responses of
a stack of local linear filters. %
Actual interaction structure is specified implicitly
by the filter coefficients.
In contrast, we learn an explicit high-order MGRF structure by considering the learning process
in terms of general exponential family distributions
nested over base models,
so that potentials added later can build on previous ones.
We relatively rapidly add
new features
by skipping over the costly optimisation of parameters.

We introduce the use of local binary patterns %
as features in MGRF texture models,
and generalise them by learning offsets to the surrounding pixels.
These prove effective as high-order features, and are fast
to compute.
Several schemes for selecting high-order features by composition or
search of a small subclass are compared.
Additionally we present a simple modification of the maximum likelihood
as a texture modelling-specific objective function
which aims to improve generalisation by local windowing of statistics.

The proposed method was experimentally evaluated by learning high-order MGRF models
for a broad selection of complex textures and then performing texture synthesis,
and succeeded on much of the continuum from stochastic through irregularly structured
to near-regular textures.
Learning interaction structure is very beneficial for textures with large-scale structure,
although those with complex irregular structure still provide difficulties.
The texture models were also quantitatively evaluated on two tasks
and found to be competitive with other works: %
grading %
of synthesised textures by a panel of observers;
and comparison
against several recent MGRF models
by evaluation
on a constrained inpainting task.

\end{abstract}

\begin{keyword}
texture synthesis and analysis; high-order MRFs; local binary patterns; structure learning.
\end{keyword}

\end{frontmatter}

\section{Introduction}

Texture modelling
is central or important to many computer vision and image processing
tasks such as image segmentation, inpainting, texture classification or synthesis,
anomaly (defect) detection, and image recognition.
Although successful specialised algorithms for texture classification,
synthesis and segmentation have been developed,
generative probabilistic models which offer relatively complete models of
statistics of individual textures are appealling.
They may be applied not only to all of the above tasks,
but to any where appearance priors or feature extraction are needed,
and they are also of interest to understanding human vision.
Generative models must capture most of the features of a texture that are significant
to human perception in order to be successful, whereas texture features used for discrimination
need not.

The most prevalent tool for image and texture modelling are Markovian undirected graphical models,
a.k.a. Markov random fields (MRFs).
An MRF together with an explicit Gibbs probability distribution (GPD)
is called herein a Markov--Gibbs random field (MGRF).
MGRFs are particularly popular for image analysis involving the determination of boundaries (as in segmentation) or
enforcing smoothness (e.g. in stereoscopic matching and image denoising).
In these cases the Markov networks are usually sparse,
with the directly interacting neighbours of each variable being
close by.
Some high-order MGRF models have been proposed for such tasks
(e.g. \cite{Ali08,Komodakis09,Wang13}),
and for binary variables efficient maximum a posteriori (MAP)
algorithms exist, such as graph cuts~\cite{Kolmogorov04}.
However things are different in the domain of
image and texture modelling,
where inference needs to be performed on real-valued or highly multivalued image variables
in dense Markov networks.
The networks used in this paper typically have Markov blankets containing 50--100
nearby and distant pixels,
and even sampling from the models proves to be difficult.

\begin{figure*}
  \centering
    \includegraphics[width=0.55\linewidth]{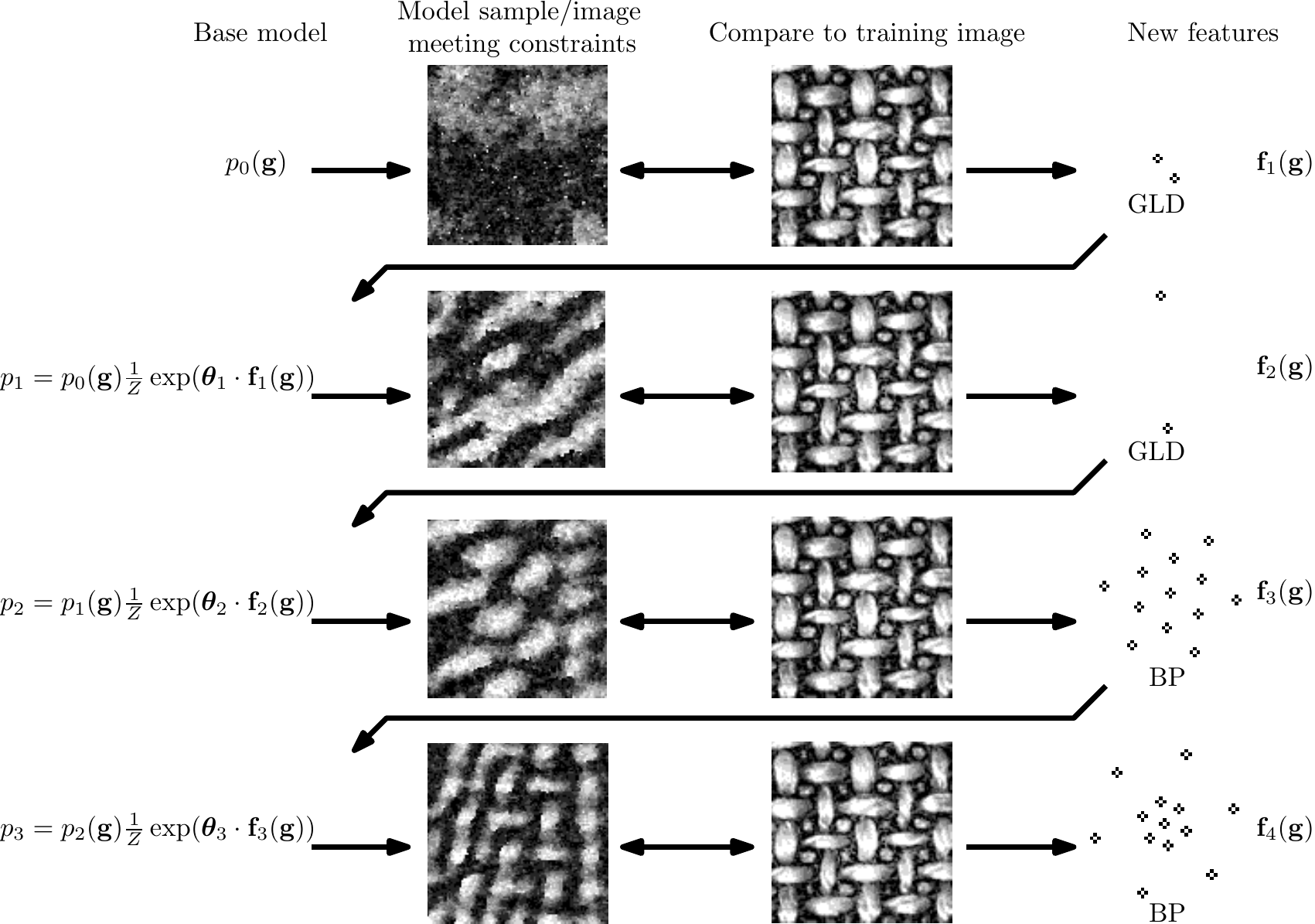}
  \caption{
    Nesting exponential-family texture models.
    At each nesting iteration images meeting the existing statistical constraints of 
    the current model are generated, such as by sampling from the model.
    One or more new features/potentials (in this example grey level differences (GLD) and binary patterns (BP))
    are selected at each iteration by searching
    for the largest deviations of their empirical marginal distributions against the target image.
    Each provides a new set of constraints, and adds a Gibbs factor to the model,
    thereby moving the model closer to the target.
  }
  \label{fig:nesting}
\end{figure*}

MGRFs and other probabilistic texture models reduce
images $\Image$ to a vector of statistics of image features $\vec{f}(\Image)$,
which are assumed sufficient to describe the texture.
The model is completed by assigning an energy $\phi$ to each feature vector,
giving a Gibbs probability distribution over images:
\begin{align}
    p(\Image) \propto \exp(- \phi(\vec{f}(\Image))).
\end{align}
Historically statistics of pairs of pixels \cite{Cross83,Geman84,Gagalowicz85}
were used.
However higher-order MGRFs (which cannot be expressed in terms of lower order ones),
have become more common as they are recognised to be necessary for more expressive
models of natural images and textures
(e.g. \cite{Komodakis09,Gimel'farb00b,Heess09,Zhu1997,FoE09,Wang13}).
Higher order interactions in image models
allow for abstracting beyond pixels, building upon larger scale image attributes like edges,
and for context and complex structures to be captured.
In addition, since regularly tiled textures have strong long range correlations
between nearby tiles
it is natural to learn an interaction structure %
(i.e. the pattern of statistic dependences between pixels)
specific to the texture. %
Yet it is still almost unheard of in computer vision and image modelling
for higher-order MRF  %
structure 
to be learned rather than hand selected.

However, selection of high-order features poses significant problems.
The cardinality of a space of possible feature functions grows combinatorially in
the order, due to both freedom in the shape of the support
(variables/pixels to select as input), %
and the need to reduce or manage its high dimensional input domain.
In other words the features should be parameterised with a reasonable number of parameters.
The higher-order MGRFs in use
nearly exclusively apply linear filters as feature functions,
with statistics of the filter responses, such as means and variances \cite{Heeger95},
correlations \cite{Portilla00},
or histograms \cite{FRAME} forming a description vector.
For texture classification many other methods of extracting useful information out of
a high dimensional pixel co-occurrence matrix have been investigated (e.g. \cite{Valkealahti98,Ojala01}).
Dimensionality can be reduced by
making assumptions such as that images are invariant to contrast and offset changes.
However this approach has seemingly been little-applied to generative texture models.

In order to tackle these problems, 
we build texture models by a model nesting procedure
which greedily selects features
and can build
higher order features by composing lower order ones.
Unlike some other works (e.g. \cite{Portilla00,FRAME})
we do not attempt to provide a fixed set of statistics/texture features to distinguish
between all textures (a goal with the Julesz conjecture~\cite{Julesz62} as its origin),
but rather learn texture-specific features.
This potentially provides compact representations
while still allowing a large and varied space of descriptors.
Each nesting iteration corrects
statistical differences between the training image and the textures class given by the previous model,
as sketched in Figure \ref{fig:nesting}.

Contributions of this paper are as follows:
(\emph{i}) We efficiently select
high-order features
by %
``nesting'' models with heterogeneous features/potentials, while coping with the difficulties
of inference in dense MGRF texture models. %
Unlike the model nesting used previously in \cite{FRAME,DellaPietra97}
we do not learn MLE parameters at each nesting iteration,
which is very expensive,
but instead generate images which match the current statistical constraints (Section \ref{sec:learning}).
These are equivalent to samples from the ideal MLE model.
Correct parameter learning can be delayed until afterwards.
We use no hidden variables as is currently popular
which eases learning and inference, %
with parameter learning remaining convex in theory.
(\emph{ii}) 
We extend the very popular local binary pattern (LBP) descriptors of images
by learning the offsets of the surrounding pixels (Section \ref{sec:selectors})
for use as high-order `binary pattern' (BP) MGRF texture features.
These are quite different from the common high-order linear filtering or Potts
potentials, and faster to compute than responses of large
linear filters.
LBPs have apparently never been used %
in this way
despite enormous popularity as image descriptors.
Experiments into texture synthesis using
MGRFs with LBP histograms as sufficient statistics
can provide insight
into the visual features actually captured by the LBPs.
(\emph{iii})
We compare several families of nested texture models utilising different high-order features,
including different methods of selecting BP offsets.
The resulting texture models have
heterogeneous feature sets composed of second-order grey level difference (GLD)
features,
and of up to 13\textsuperscript{th}-order BP features
or Laplacian of Gaussian and Gabor filters.
The use of learned long range GLD interactions allows almost-regular (tiled)
textures in particular to usually be synthesised well.
(\emph{iv})
The ability of the proposed procedure
to learn characteristic features across different
types of textures is
demonstrated with texture synthesis across a varied set of greyscale textures,
also evaluated by a panel of observers,
along with several other comparisons.
(\emph{v})
In order to improve generalisation and to attempt to allow partially inhomogeneous training images
a variant on the maximum likelihood estimate (MLE) %
was used, such that the training image is split into pieces and the minimum of the
likelihoods of the pieces rather than their product is maximised.

\section{Related Work}
\label{sec:relatedwork}

\subsection{High-order MGRF models}

Much research in texture analysis has focused on describing textures using the distributions
of responses of square linear filters.
MGRF texture models utilising non-trivial filters were introduced with FRAME \cite{FRAME,Zhu1997},
where the filters were selected from a manually-specified bank.
More recently learning the filters themselves (with predetermined fixed supports)
has been popularised
by the Field-of-Experts (FoE) model~\cite{FoE09}.
These model the marginal distibution of responses for each filter with hand-picked potentials with few parameters;
in the original FoE model they are student-t distributions.
Therefore the interaction structure is learned as filter coefficients.
FoE was extended to bimodal FoE (BiFoE) which uses more informative bimodal potentials,
and successfully applied to texture modelling~{by Heess et al.}~\cite{Heess09};
several state of the art generative texture models
have been built on BiFoE,
some using various configurations of hidden variables.
Kivinen and Williams~\cite{Kivinen12} improved on BiFoE by using
gated MRFs \cite{Ranzato10},
and Luo et al.~\cite{Luo13} investigated convolutional deep belief networks (DBN)
and spike-and-slab potential functions. %
However because of learning difficulties and to reduce required computation
all these learned-filter models have been restricted
to relatively small filter sizes. These do not directly capture distant interactions;
the largest filter size used in the mentioned works was $11 \times 11$.
As a result, while these filter-based MGRFs
model certain classes of textures excellently, they have inherent weaknesses.
A more general survey of MRFs which covers high order models can be found in~\cite{Wang13}.

The recently popular hierarchical filter-based MRFs
are a merger of MRF models, especially Boltzmann machines, and
models of single image patches using independent components analysis \cite{Hyvarinen00}
and overcomplete Product-of-Experts \cite{PoE03}.
These are applied to larger images by tiling and overlapping the filters.
Today MRFs including latent variables are popular,
often inspired by simple and complex cells in the visual cortex,
with strong links to
convolutional artificial neural networks.
Using multiple layers allows application to high level computer vision tasks
where the Markov property does not hold,  such as object recognition
and face modelling.
In many cases integrating out the latent variables leads to a completely
connected, non-Markovian interaction graph.
We diverge from this direction to consider
the direct inclusion of higher-order features in models,
which is more common for other machine learning tasks
where there often is
no analogue to linear filtering.

\subsection{Texture features}
Various high-order local texture descriptors have %
been applied to texture classification (e.g. \cite{Valkealahti98,Ojala01,Sharma12}) going back at least 20 years.
For such applications, invariances to contrast, offset, scale, rotation,
and deformation are given particular weight.
Non-linear texture features have also
been used in MGRF texture models; Sivakumar and Goutsias~\cite{Sivakumar99}
introduced MGRF texture models which used sophisticated multi-scale features defined through
mathematical morphology.
Many sought to directly
reduce the dimensionality of high-order co-occurrence histograms
through traditional techniques such as
spectral clustering of 
histogram bins \cite{LiaoYoung10},
vector quantisation \cite{Ojala01},
Gaussian mixture models %
\cite{Sharma12}, %
and self-organising maps \cite{Valkealahti98}.
However such dimensionality reduction techniques usually require
a nearest-neighbour search to map new data points, which likely makes them unsuitably
slow as feature functions in MGRFs.

Probably the most popular texture descriptors for
discrimination
are LBPs~\cite{Ojala96},
which compare the intensities of a ring of pixels $p_1, \ldots, p_k$ around a central one $p_0$,
and form the bit-vector $(p_i > p_0)_{1 \leq i \leq k}$.
These are nearly contrast- and offset-invariant and can
be straight-forwardly extended to rotation invariance \cite{Ojala02}
by merging bins, and extended to partial scale invariance by using multiple concentric rings.
In addition to their proven ability to distinguish textures
LBPs are also very cheap to compute,
hence we investigate their use in texture modelling.
Nosaka et al. \cite{Nosaka13} introduced co-occurrence statistics
of neighbouring LBPs in a way that is rotationally invariant,
This is particularly interesting for future extension of the BP-based MGRFs in this
paper to rotational invariance.
LBPs have inspired a number of other variants such as
local ternary patterns (LTPs) \cite{TanTriggs07},
and local radial index \cite{Zhai13}.
Liu et al.~\cite{Liu13:2,Liu14:1}
have also used LBPs, LTPs, and related non-linear ``ordinal'' texture descriptors with learnt offsets, and applied them to texture classification and retrieval.
Although explained within the framework of MGRFs, these approaches do not involve learning
a complete MGRF model.
They built up the higher-order cliques (up to 20\textsuperscript{th }order) out of 
the lower-order ones by searching
for cliques in an interaction graph computed using a thresholding rule.
We find this too restrictive and pursue alternative strategies for selecting the offsets.

\subsection{Structure learning and nesting}

A method for selecting the features of the
model which is both intuitive and theoretically sound
is to use greedy \emph{sequential structure selection},
as has been used by various authors
\cite{DellaPietra97,Zalesny99,FRAME,Grafting,GraftingLight}
with a number of details varying.
This alternates between adding one or more features/factors to the current model from
a candidate set according to an estimate of the best feature to add,
and then finding the new MLE of the
parameters (initialised at those from the previous iteration).
The most common metric for selecting the best feature  %
to introduce is to %
select that with the largest `error' between training image and model
synthesis result
\cite{Zalesny99,FRAME,Grafting}.
This paper follows the same general approach, which we call `nesting',
although we describe our particular flavour.
The main difference in our approach
is to attempt to directly generate images matching the statistics at each iteration,
learning approximate parameters as a side effect.

Della Pietra et al. \cite{DellaPietra97} treated structure learning
in MGRFs (for natural language processing  problems) as a feature selection problem,
and built up higher order features out of lower order ones.
Several other authors have since considered selecting MGRFs features by
gradually composing together low order atomic features (`general-to-specific')
(e.g. \cite{Schmidt10,McCallum03}),
or by starting from template-like features (sometimes called `patterns')
which are conjunctions of simple $x_i=c_j$ predicates, derived directly from the
training data and gradually generalising them (`specific-to-general') (e.g. \cite{Mihalkova07}).

A closely related popular method for MGRF structure learning is the use of
sparsity-inducing $L_1$ regularisation \cite{Grafting,LGK06}, which forces some feature weights
to exactly zero so that they can be removed.
Otherwise proceeding like nesting, this approach
has the advantages that the regularisation combats overfitting,
that it allows removing a feature after it has
been added, and is a convex problem.
Recently Chen and Welling~\cite{Chen12} suggested the use of spike-and-slab
priors (similar to $L_0$ regularisation) instead of $L_1$ regularisation.
This has some advantages, but does not also lead to an efficient greedy
algorithm with an optimality guarantee.

\subsection{Texture synthesis}
Practical texture synthesis is currently %
dominated by algorithms which combine pieces of a source image
so that the pieces fit together well.
This is defined in terms of the match between neighbourhoods of the pieces.
Region-growing techniques add one pixel (e.g. \cite{Efros99,Wei00})
or patch (e.g. \cite{Efros01,Kwatra03}) to the image at a time.
Algorithms are often multiscale \cite{Popat93,Paget98,Wei00}.
However all these synthesis algorithms have the property
that they copy large parts of the source image directly
into the result, either an explicit part of the algorithm,
or an emergent property of the algorithms'
search for best matching neighbourhoods,
due to `forced moves'.

Another rather successful but not realtime class
of texture synthesis algorithm is based on global optimisation
of the image
such as by projection onto the set of images with certain statistics
equal to those of the training image,
in particular statistics of wavelet-decompositions \cite{Heeger95,Portilla00}.
A codebook of texture patches or primitives can also be used instead~
\cite{Peyre09}.
These are implicitly MRF texture models
as they provide no explicit probability distributions
(hence our discrimination between MGRFs and MRFs)
and so are less broadly applicable than probabilistic models.
Hence, although these give good synthesis results
these algorithms in fact attack a different problem
than the more statistical one that this paper does;
we use texture synthesis only for evaluation.

\section{Nested Markov--Gibbs Random Fields}
\label{sec:model}

This section provides first some definitions, and then defines
general exponential distributions as the conceptual and theoretical basis
for MGRF model nesting.
A description of the nesting algorithm follows.

\subsection{Basic notation and definitions}

Any probability distribution which is nowhere-zero can be represented
as a GPD factorised into Gibbs factors: functions of complete subgraphs (cliques)
of the Markov network.
In this paper we consider the class of MGRFs with
factors (synonymously, potentials) that can be described as the product of
a fixed feature function
---identifying a certain signal configuration (pattern)---
and a corresponding weight/parameter.

We restrict our scope to modelling of homogeneous textures
by repeating
the Gibbs factors and cliques across the image to achieve
translation invariance.
Let $\Grid \subset \Z^2$, where $\Z$ are the integers,
be a finite  lattice of image coordinates
and
$\boldsymbol{\alpha}=
\left\{\alpha_i: \alpha_i \in \Z^2; 1 \leq i \leq d\right\}$
be a list of $d$ coordinate offsets with $\alpha_1 =(0,0)$ fixed.
An order $d$ clique family $C_\Clique$ in $\Grid$
is the set of all spatially repeated cliques
with the offset pattern given by $\Clique$:
\begin{align*}
  C_\Clique :=  \{ ( r_1, \ldots, r_d ) :
                 r_1, \ldots, r_d \in \Grid; \
				  r_i-r_1 = \alpha_i;\ i=1,\ldots,d \}.
\end{align*}
Let $\Image : \Grid \to \{0, \ldots, Q-1\}$ be an image on $\Grid$ with $Q$ possible
grey levels.
An order $d$ feature
is a  function $f : \{0, \ldots, Q-1\}^d \to \N_s$ with finite range $\N_s := \{1, \ldots, s\}$
and an associated clique family $C_{f}$ given by offset list $\Clique_{f}$.
The histogram of values of $f$ collected over an image $\Image$ is the vector
\begin{align}
  \vec{h}_f(\Image)
    &= \left[ \sum_{c \in C_f} \llbracket l = f(\Image_c) \rrbracket \ : \ l \in \N_s \right]
\end{align}
where $\llbracket \cdot \rrbracket$ is the Iverson bracket mapping true $\mapsto1$, false $\mapsto0$, and $\Image_c$ is
notation for
sequence of pixels $\Image(r_1), \ldots, \Image(r_d)$, $r_i \in c$, $c$ a clique.
The order of a model is defined as the maximum order of any of its features.

\subsection{General exponential distributions}
\label{sec:general-exp}

In the nesting procedure, new features to be added to the current model are selected based on the
disagreement between their statistics in the training image and
the model's expected statistics, approximated from
model samples.
In this way the model is modified to correct these errors.
When the statistics in question are expectations (equivalently, histograms), these
corrections take a particularly simple form.

Let $F_{i+1}$ be a set of feature functions %
and let the histogram vector
 $\vec{f}_{i+1}(\Image)$ be the concatenation of histograms for the
set $F_{i+1}$: $\vec{f}_{i+1}(\Image) := [\vec{h}_f(\Image) : f \in F_{i+1}]$.
A general exponential family model  \cite{Jaynes57}
is a probability distribution which can be written in the following
form specified by a base model $p_{i}(\Image)$,
a vector-valued feature function $\vec{f}_{i+1}(\Image)$,
and a parameter vector $\params_{i+1}$,
\begin{align}
    p_{i+1}(\Image | \params_{i+1}) = \frac{1}{Z(\params_{i+1})} p_{i}(\Image) \exp(- \vec{f}_{i+1}(\Image) \cdot \params_{i+1})
    \label{eq:nested-MGRF}
\end{align}
where $Z(\params)$ is a normalisation factor, and $\cdot$ denotes dot-product.
In our case, the parameters are a concatenation of per-feature parameter vectors $\params_{i+1} = [\params_f : f \in F_{i+1}]$.

If one wishes to find a model $p_{i+1}$
meeting constraints to be satisfied
in the form of expectations
$\mathbb{E}_{p_{i+1}}[\vec{f}_{i+1}(\Image)] = \vec{f}_{i+1}(\ImageObs)$
(where $\ImageObs$ is given training data),
but already has prior information expressed as a
base model (in this case the model at the previous iteration,  $p_{i}$),
then it is widely known \cite{DellaPietra97} that the probability distribution $p_{i+1}$
which matches the new constraints but deviates from the base
model the minimum possible amount (as measured by the Kullback-Leibler divergence,
i.e. has maximum entropy relative to $p_i$)
has a simple form.
It is the general exponential family model
given in \eqref{eq:nested-MGRF}
with
the maximum likelihood estimate (MLE)
of the parameters $\params^*_{i+1} := \arg\max_\params p_{i+1}(\ImageObs|\params)$.
It can be seen from Eq. \eqref{eq:gradient}
that the MLE parameters achieve the desired statistical constraints.
Technically, the MLE $\params^*_{i+1}$
may lie at an infinitely distant point in parameter space,
which is not permitted by the usual definition of a GPD.
This possibility is easily avoided by slightly smoothing $\vec{f}_{i+1}(\ImageObs)$.

\subsection{Model nesting}
\label{sec:search}

The challenges of both selecting high-order features and structure
can be met by gradually building up features from pieces,
reducing the set
of candidate features to a size that can be exhaustively searched at each iteration.
One could reasonably expect that when
some configuration of pixels
(as recognised by a feature function) is characteristically common for a texture,
that the configuration
restricted to a subset of pixels would usually also
be common.
Hence we conjecture that in practice high-order feature functions
picking out characteristic interactions
can be found by building up from lower order features.

From the theory of general exponential distributions
we can see that there is no need to modify whatever parameters the
base model may have, although
in most previous works involving sequential structure learning this is normally done
as it allows increasing the likelihood.
However to attempt to reduce overfitting we suggest that
it is reasonable to only adjust parameters associated
with the new features $\vec{f}_{i+1}$.
Dud\'{i}k et al. \cite{Dudik07} described %
such a coordinate-wise descent
algorithm for a wide class of generalised maximum-entropy
problems which
selects a best single parameter to update on each iteration.
They state the standard approach of updating all parameters each iteration
"is impractical when the number of features is very large."
However in our experiments we found that this was too aggressive
a solution to overfitting, and obtained better results
by keeping all parameters free.

Given a set of feature functions $F_{i}$, having already been selected,
a \emph{selector} function $C(F_{i})$ is defined 
to provide a candidate set of new features,
possibly built upon previous features.
The algorithm proceeds stage-wise
through a sequence of selectors $C^1, \ldots, C^k$
each providing features of a certain order and type,
to avoid the problem of comparing across feature types.
Sequential selection simply selects one or more feature functions among 
$C^j(F_i)$ on which the current model $p_i$ has the largest disagreement (error)
against the training data
$e(f) := d(\vec{h}_f(\ImageObs), \vec{h}_f(\Image^{(i)}))$,
where $d$ is a distance function on histograms.
The next selector in the sequence is proceeded to
after $\max_{f\in C^j(F_i)} d(\vec{h}_f(\ImageObs), \vec{h}_f(\Image^{(i)}))$
becomes too small, and the algorithm terminates when the
selectors are exhausted.

One possible variant is to use separate training and validation
datasets, and to stop when performance (measured either by $d$
or an application-specific external evaluation function such as texture
discrimination accuracy) on the validation begins to decrease.
Such a validation test was used in the earlier work \cite{Versteegen14},
in order to detect slightly non-homogeneous textures.
This paper takes the idea further that statistical variations
across the training data are important:
Section \ref{sec:minimax} discusses the selection of
features from a set of pieces of the training images rather than
less powerfully only
looking for variation between a single training and validation pair.

Previous  sequential feature
selection implementations 
have used $\ell_1$~\cite{FRAME,LGK06} or $\ell_2$ distance
or `gain' for $d$.
Gain is the increase in information 
contained in a model
(equivalently, decrease in entropy) due to adding a feature to it,
and can be either be estimated analytically \cite{Zhu1997} or,
at great expense,
by actually comparing every possible extended model \cite{DellaPietra97}.
Zhu et al. \cite{Zhu1997} also gave an correction to the gain
for theoretical uncertainty in the estimated statistics,
which is a form of the Akaike information criterion (AIC),
penalising model complexity.

\begin{algorithm}[htb]
  
\begin{algorithmic}
\REQUIRE Initial base MGRF model $p_0$ with feature set $F_0$,
  training image $\ImageObs$
\STATE{$i \gets 0$}
\FOR{each selector $C$ in $C^1, \ldots, C^k$}

  \LOOP
    \STATE{Obtain set of samples $\{\Image_l\}_{1 \le l \le n}$ from model $p_i$}
    \FOR{each $f \in C(F_i)$}
    \STATE Collect histograms $\Hist{f}(\ImageObs)$ and $\Hist{f}(\Image_1), \ldots, \Hist{f}(\Image_n)$
    \STATE $\bar{\mathbf{h}}_f(\bar{\Image}) \gets \frac{1}{n}\sum_l \vec{h}_f(\Image_l)$
    \ENDFOR
    \STATE {Pick one or more $f$ with maximal
      $d(\bar{\mathbf{h}}_f(\bar{\Image}), \Hist{f}(\ImageObs))$ for some histogram distance measure $d$

    \STATE{ $F_{i+1} \gets F_i \cup \{ f \}$}
    \STATE{
      Form  $p_{i+1}$ by adding $f$ to $p_i$
      as in Eq. \eqref{eq:nested-MGRF}}
    }
    \STATE{
     Learn parameters $\params_{i+1}$ for the new features}

    \STATE{ $i \gets i+1$ }
  \ENDLOOP
  \ENDFOR
\end{algorithmic}
  \caption{Outline of basic MGRF nesting procedure}
  \label{algrthm:nesting}
  
\end{algorithm}

However, real uncertainty in estimates found through
Markov chain Monte Carlo (MCMC) sampling
from a MGRF is typically vastly larger than theoretical estimates of
uncertainty because the sampling
may not converge (see Section \ref{sec:sampling}).
This is especially true
because the quality, and also statistics, of
the approximate samples that we attempt to rapidly draw from the current texture model
are highly variable.

We used the Jensen--Shannon divergence (JSD) \cite{Lin91} 
for distance function $d$ %
as a proxy to the additional information content of  $\vec{h}_f(\ImageObs)$.
The JSD is defined as
\begin{align*}
  D_\JS(p \parallel q) &:= \frac{1}{2} \left(
  D_\KL \left( p \parallel m \right) +
  D_\KL \left( q \parallel m \right)
  \right)
\end{align*}
where $D_\KL$ is the Kullback-Leibler divergence (KLD), $p$ and $q$
are probability distributions, and 
$m$ is the averaged distribution $\frac{1}{2}(p + q)$.
The JSD measures the discriminability of two distributions,
namely the average certainty (as measured in bits) with which a sample
can be ascribed to one or the other.
The JSD is symmetric, bounded in
the range $[0,1]$ bits (when computed using base 2 logarithms)
and is popular because it is suitable for comparing two
empirically estimated distributions, unlike the KLD, which is often undefined.
In practice, the JSD usually provides very similar rankings to the $\ell_1$ distance.

\begin{figure}[htb]

  \centering
      \includegraphics[width=0.119\linewidth]{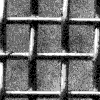}
      \ \includegraphics[width=0.119\linewidth]{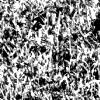}
      \ \includegraphics[width=0.119\linewidth]{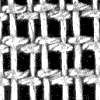} 
      \ \includegraphics[width=0.119\linewidth]{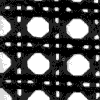}
      \ \includegraphics[width=0.119\linewidth]{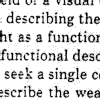}\\ \vspace*{3mm}
       \includegraphics[width=\linewidth]{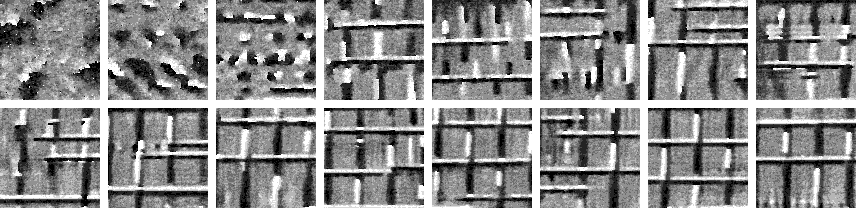}\\ \vspace*{2mm}
       \includegraphics[width=\linewidth]{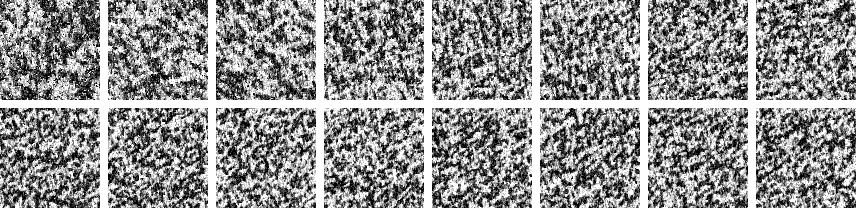}\\ \vspace*{2mm}
       \includegraphics[width=\linewidth]{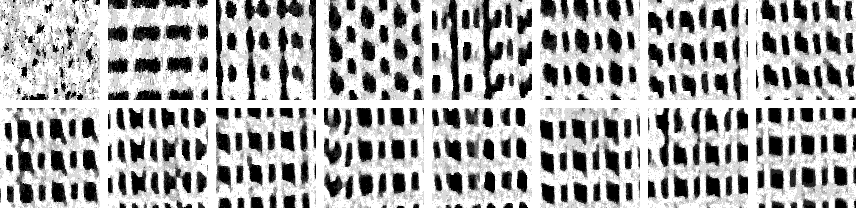}\\ \vspace*{2mm}
       \includegraphics[width=\linewidth]{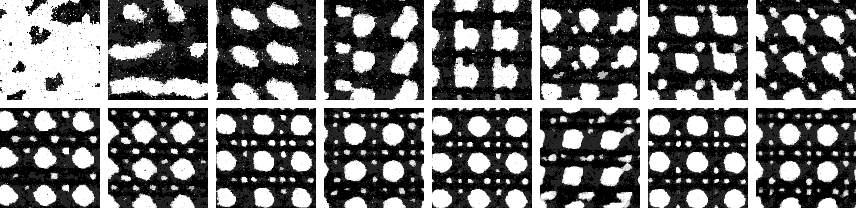}\\ \vspace*{2mm}
       \includegraphics[width=\linewidth]{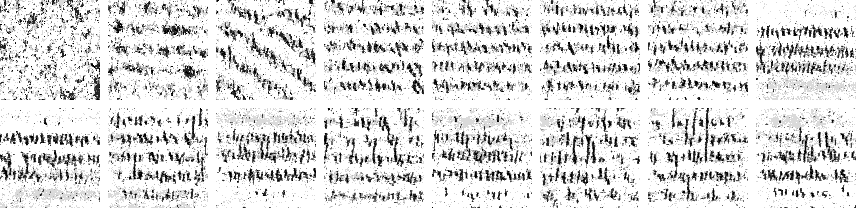}\\ \vspace*{2mm}

  \caption{
    Examples of the nesting procedure. The top row shows pieces of the original five
    textures (from left to right: D1, D9, D20 and D101~\cite{Brodatz} and `text'~\cite{Portilla00}). The subsequent five pairs of the rows 
    illustrate the progress of feature selection. 
    Each image (from left to right) shows one of the CSA `samples' at a nesting iteration,
    which approximately matches the statistical constraints of the current model (see Section \ref{sec:learning}).
    The first row of each pair shows addition of pairwise GLD features, and the second shows 
    addition of the 9\textsuperscript{th}-order
    jagged star features (see Section \ref{sec:selectors}).
  }
  \label{fig:nesting_ex}
\end{figure}

\begin{figure}[htb]
\centering
      \includegraphics[width=\linewidth]{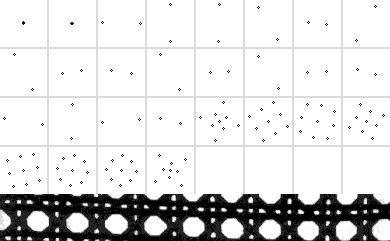}\\ \vspace*{1mm}
     \includegraphics[width=\linewidth]{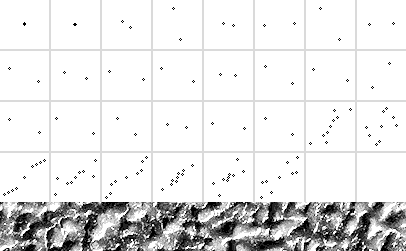}

    \caption{
      Shapes of the 2\textsuperscript{nd}- and 9\textsuperscript{th}-order `jaggered star' cliques selected 
      for D101 and Metal0004~\cite{VisTex-webpage}, in order of selection.
      A piece of the texture at the same scale is shown at bottom.
    }
    \label{fig:cliques1}
\end{figure}

Algorithm \ref{algrthm:nesting} summarises this approach.
Figure \ref{fig:nesting_ex} gives examples of its operation,
starting from the initial model (from which the first sample is drawn) which included
only marginal and nearest-neighbour pairwise potentials and proceeding \\
through two selectors.
The first selector returns three pairwise potentials at a time.
Frequently, there is no visible improvement from one step to another,
however even the samples getting worse do not indicate that the
model is getting worse, due to the unstable sampling process.
There is also usually a big jump as each new class of features is added,
and decreasing returns thereafter.
Determining a suitable stopping point for model learning is difficult, as the histogram
distances  $e(f)$ do not reflect visual proximity to the training image very well,
and a texture similarity metric such as
improved structural texture similarity (STSIM2) \cite{Zujovic11}
might be a better option.
For experiments in this paper we instead used a fixed 8 iterations
for each selector, as this removes the variability due to an imperfect
stopping rule, which can have a large effect.
Figure \ref{fig:cliques1} shows the cliques selected in
two of the examples in detail.

\section{Texture Modelling}
\label{sec:texture-modelling}

The procedure described above is generic and could equally
well be applied to learning MGRF models for machine learning tasks
outside of computer vision.
Its refinement specific to texture-modelling,
including some of our implementation details is given below.

Broadly, there are two ways to achieve desired invariances:
by designing the models to enforce those invariances (e.g. by using
feature functions with some invariance property),
or by learning them from suitable training data which demonstrates them.
We use both approaches, although the latter
is only preliminarily developed here.

\subsection{Texture-specific maximisation of the minimum likelihood}
\label{sec:minimax}

Up to now, we have described
model learning in terms of MLE that
maximises the total probability of the training image,
but the sufficient statistics of the image are the mean
values of the feature functions over the image.
This is completely oblivious to variations across the image,
and for this reason some small parts of the training image may
actually have low probability. %
This is particularly true of textures composed of large randomly
distributed textons (coherent texture elements like pebbles or grass blades),
which greatly reduces the effective amount of training data,
and thus is the most difficult to learn for all kinds of MGRF models.
At one place in the image or over the whole image a particular angle or offset between
textons may be more common than the average just by chance.
Just like humans, sequential feature selection easily sees patterns in noise.
We have already made the assumptions that the texture is homogeneous with
the Markovian property
and that the entire training image is a sample from the
texture class, without foreign inclusions.
We can thus expect that every piece of the training image of sufficient
size (which we term the ``scale'' of the texture) is itself also a visually recognisable sample of the texture
class, while those much smaller aren't,
and thus the model should also recognise them.
The scale will be approximately equal to once to twice the tesselation offset if the texture is regular,
related to the size of any textons that the texture is composed of,
and in general similar to the spatial extent of the Markov blanket.
Indeed, this property holds for all of the training images
used in our experiments.
Hence it is unsatisfactory that only averages over the training image are considered,
and we propose to switch away from the MLE as our objective.

Let $\PiecesObs$ be a set of training images.
In practice we simply split a single training image $\ImageObs$ into pieces
of size $80 \times 80$ with overlap of $22$ pixels.
This is related to the maximum offset length between interacting pixels that we search for, $40$ pixels.
We change our optimisation problem to
\begin{align}
  \max_{p,\params} \min_{\ImageObs_i} \ell(\params|\ImageObs_i). \label{eq:objective2}
\end{align}
It is perhaps desirable to use a smoothed differentiable version of the objective using a `soft-min' instead of min,
\begin{align}
  \max_{p,\params} \sum_{i} \ell(\params|\ImageObs_i) w(\ImageObs_i\ |\ p(\cdot|\params) ),
\end{align}
where $w$  weights the pieces exponentially according to their proximity to the minimum
(which is easy to compute):
\begin{align}
\begin{aligned}
w(\ImageObs_i | p ) &:=& \frac{ \exp ( -\alpha \ell(\params|\ImageObs_i) ) } { \sum_{j} \exp ( -\alpha \ell(\params|\ImageObs_j) ) }  \\
   &=& \frac{ \exp ( -\alpha \params \cdot \Hist{}(\ImageObs_i) ) } { \sum_{j} \exp ( -\alpha \params \cdot \Hist{}(\ImageObs_j) ) } 
\end{aligned}
\end{align}
where $\alpha$ is a softness parameter.

This objective implies that when selecting features we replace the criterion
\[
\arg \max_f d(\bar{\Hist{f}}(\ImageSample), \Hist{f}(\ImageObs))
\]
with one of
\begin{align}
  \arg \max_f \min_i d(\bar{\Hist{f}}(\ImageSample), \Hist{f}(\ImageObs_i))
\end{align}
or
\begin{align}
  \arg \max_f \sum_i w(\ImageSample|p) d(\bar{\Hist{f}}(\ImageSample), \Hist{f}(\ImageObs_i))
\end{align}
to find features that are characteristic for all parts of the image.

\subsection{Second order methods} %
\label{sec:gradient}

As the data likelihood in the base model $p_i(\ImageObs)$ is fixed w.r.t. $\params$ and can be dropped,
parameters of general exponential family distributions are learned in
the same was as without a base model. %
The gradient vector $\boldsymbol{\nabla}\ell(\params|\ImageObs) $
and the Hessian matrix
$\mathbf{H}(\ell)(\params|\ImageObs)$
of the log-likelihood are easily derived as: 
\begin{align} %
\boldsymbol{\nabla}\ell(\params|\ImageObs)
&  = 
-\Hist{}(\ImageObs) + E_\params \left\{\Hist{}(\Image)\right\}  \label{eq:gradient} \\
\mathbf{H}(\ell)(\params|\ImageObs)
& = -\textup{Cov}_{\params}\{\Hist{}(\Image)\}  \label{eq:hessian}
\end{align}%
 where $E_\params\{\Hist{}(\Image)\}$ and $\textup{Cov}_\params\{\Hist{}(\Image)\}$
denote respectively the expected value and covariance matrix of the image features
$\Hist{}(\Image)$ w.r.t. the distribution $p(\Image|\params)$.
As the covariance matrix is always non-positive definite, the log-likelihood is
concave down
in the space of parameter vectors, which is a common property of all exponential family
distributions \cite{Barndorff82}.

Computing the expectation in Eq. \eqref{eq:gradient} exactly is in general intractable.
Its approximation using a Markov chain Monte Carlo (MCMC)
sampler such as the single-site Gibbs sampler
can be highly unreliable,
because the energy landscape often has deep local minima
which are inescapable in reasonable time-scales.

The Hessian may be reasonably and easily approximated as a diagonal matrix,
so given samples from the model $p(\Image|\params_j)$ we can
find the 2nd order Taylor expansion about $\params_j$
as an approximation to the log-likelihood
using Eq.s \eqref{eq:gradient} and \eqref{eq:hessian}.
A Newton step can be performed by inverting the Hessian,
while a more numerically stable alternative is to find the maximum
of the 2nd order likelihood approximation along the direction of the gradient.
See \cite{Gimel'farbZhou08} for details.

We use this second-order method to find an initial
approximation to the parameters at each nesting iteration,
to save a little time.
As both gradient and Hessian are highly approximate
and Newton's method takes dangerously
large steps we take a single 2nd order step,
and then switch to a stochastic 1st order method which follows only the approximate gradient,
as detailed in the next section. %

\subsection{Controllable Simulated Annealing}
\label{sec:sampling}
\label{sec:learning}

Assuming that the time to sample from a MGRF is linear in the number of
features, the nesting procedure runs in time quadratic in the number of
nesting iterations.
In order to keep running times reasonable, the number of nesting
steps should be limited, and computation per nesting iteration minimised.

The most obvious  method to learn the parameters of a MGRF
is to perform stochastic gradient descent (SGD),
following a noisy approximation to the gradient
given by sampling from the model to approximate
the expectation in the gradient (Eq. \eqref{eq:gradient}).
Sampling could be performed by starting from an initial
image (e.g. of noise) and performing Gibbs sampler
sweeps over the whole image until its total energy converges.
However, this method is very slow.
The MGRFs we learn are very dense, which makes sampling problematic;
Gibbs sampling may mix very slowly, and easily gets
stuck in local minima.   %
Alternatively, a fixed number of sweeps can be performed
to produce a sample,
which is still useful for synthesising images but optimises
the wrong objective.

Rather than repeatedly sampling from the models
to approximate the gradient
we used controllable simulated annealling (CSA) \cite{Gimel'farb99},
simultaneously invented in \cite{FRAME},
to produce approximate images matching the training statistics
much faster than possible learning MLE parameters.
An essentially identical procedure was
also reinvented in~\cite{Tieleman08} under the name `persistent contrastive divergence' (PCD).
The goal in PCD is parameter learning, and it is presently the most popular method for that
for MGRFs.
The difference is that PCD
uses small stepsizes, and is started from the training
data rather than a random image.

CSA alternates Gibbs sampling sweeps (visiting each pixel
in the image $\ImageSample$ once) and changes to parameters according to
\[
 \Delta\params = \boldsymbol{\lambda}_t \circ (\Hist{}(\ImageSample) - \Hist{}(\ImageObs))
\]
where $\circ$ denotes element-wise Hadamard product, $\ImageObs$ is the training image, $\boldsymbol{\lambda}_t$ is a vector of the current
per-parameter step sizes and $t$ is the iteration number.
A Robbins-Monro (RM) sequence $\boldsymbol{\lambda}_t = \frac{15}{15 + t} \vec{1}$ was used,
where $\vec{1}$ is the all-ones vector of the same dimension as $\params$.
CSA can more rapidly produce an image with statistics which usually match the
training statistics fairly well, but also has a strong tendency to oscillate.
For this reason we introduce a variant of CSA
called accelerated controllable simulated annealing (ACSA),
which uses an
gain vector adaptive step size instead of RM steps, as described
by Almeida et al. \cite{Almeida99}, with
default parameters and initial step size vector $\boldsymbol{\lambda}_0 = \vec{1}$.
This uses different learning rates for different parameters,
dampening oscillations by reducing the learning rate for parameters which do so,
while increasing the learning rate for those that seem to require it.
Adaptive step sizes proved to give far faster convergence
to the desired statistics than RM steps,
and are more robust to initial step size.

Finding optimal parameters which produce good unguided synthesis results
requires far more fine tuning to eliminate unwanted energy minima
which are approached very slowly during MCMC.
In theory, if optimal parameters are learned then
MCMC sampling and CSA should both produce
samples which have statistics equal to the constraints on average,
and are indistinguishable by the model features.
Hence CSA `samples' can be substituted for real samples
from the MLE model.
Ideally if available a more efficient sampling algorithm than single-site
Gibbs sampling should be used, but most popular samplers are not applicable here.
More efficient algorithms have been developed for creating
images matching desired statistics
by making modifications attempting to directly move towards them \cite{Heeger95,Portilla00,Zhu00}.

\subsection{Sampling in practice}

Naturally, CSA will converge to the desired statistics faster if starting
from parameters close to the MLE.
At each nesting iteration
we first found an approximation to the parameters as described in Section \ref{sec:gradient},
and then ran CSA four times for 50 Gibbs sweeps each, starting from uniform noise images, to produce four samples of size $100\times100$.
CSA results can be noisy and unreliable, not reaching the desired statistics,
so it helps greatly to average over multiple runs.
The parameters produced as a side effect of CSA
are very approximate and may actually diverge from the MLE,
but nonetheless are useful and carried forward to future iterations.

An important consideration when drawing samples from a MGRF texture
model using an MCMC sampler is the choice of initial image.
While the Markov chain converges asymptotically  to the
model distribution, this can take a number of steps exponential
in the number of pixels
due to ubiquitous deep local minima.
However it is well known that MCMC converges slowly when highly correlated
variables are updated independently, and in image models all nearby
pixels are typically highly correlated to one another, making Gibbs
sampling inefficient and prone to being trapped in local minima.
The most common example of these are `crystallisation'
of regular patterns growing from two or more areas and meeting
without aligning.
Gibbs sampling of regular textures starting from noise often
leads to crystallisation if the image size is several multiples
of the maximum clique size.

There are a number of ways to avoid crystallisation.
One is to add longer range interactions in order to more strongly enforce
global structure.
Otherwise, the process can be pushed towards the correct regularity
by starting from an initial image with a template, in the form of
dots, stripes, or any other unevenness of the desired tesselation.
Such tesselation of near-regular textures can automatically be extracted
by computing co-occurrence statistics for different offsets,
in the form of a model-based interaction map (MBIM) \cite{Gimel'farbZhou08}.
Other alternatives are to slowly grow the image by extending the boundaries,
starting from a single small `seed' which is first allowed to converge \cite{Versteegen14},
or starting from a random piece of the training image as a seed.
The latter was used for these experiments due to simplicity.
If a texture model is to be used for other tasks such as segmentation or classification
better performance could be expected if appropriate
learning, sampling and validation methods are used.
For example
for texture synthesis we care only about the local mimimum that
is reached on sampling from the starting sample (a white noise image
in this work), and disregard the energy landscape outside of this basin,
while for texture classification
we want to ensure correct behaviour for distant images too.

For unguided texture synthesis after learning a model we used
300 sweeps of ACSA,
growing from a seed, to produce images of size $180\times180$ plus trimmed margins of up to $65$
pixels depending on feature sizes.
Hence when inpainting this shortcut is not available:
in practice CSA and ACSA
are harmful for inpainting as generalisation (e.g.
differing lighting) may be required.
Instead we tuned the model parameters in the normal way,
and then used Gibbs sampling to inpaint.

\subsection{Binary patterns}

In order to reduce the space of candidate potentials, we define
families of potentials that are parameterised only by the shape of
their supports, with unparameterised feature functions.
The simplest such order $k$ feature on an image of $Q$ pixel grey-levels is
the trivial grey level co-occurrence (GLC) feature
\[
f_{\textup{GLC}}(x_1,\ldots,x_k) := x_1Q^0 + \ldots + x_k Q^{k-1}
\]
with $Q^k$ bins.
GLC features proved to have poor generalisation ability;
while they performed well for texture synthesis, when
used for inpainting tasks
they remembered the original contrast and offset of the training image
rather than matching the contrast of the surrounding inpainting frame.
Previous researchers have found that histograms of the pairwise grey level difference (GLD)
--- defined for $k=2$ as $ f_{\textup{GLD}}(x_1, x_2) := x_2 - x_1,$
with $2Q-1$ bins ---
encodes the large majority of the information in a pairwise GLC histogram \cite{Unser86,Ojala01},
confirmed by our own comparisons GLC and GLD features for texture synthesis (see Section \ref{sec:synth}).
Hence we used GLD instead of GLC features to capture pairwise interactions.

The high-order features investigated were binary patterns (BPs),
generalising LBPs
by using learned offsets from the central pixel.
The BP feature function thresholds the grey levels
of the pixels in each clique against a certain distinguished
pixel (no longer necessarily in the centre):
$f_\textup{BP}(\Imagepix_{0}, \ldots, \Imagepix_{d}) := \sum_{1\le i \le d} 2^{i-1}\llbracket\Imagepix_{0} < \Imagepix_{i}\rrbracket$.
No interpolation between pixels was performed,
nor were histogram bins combined as in
uniform LBPs \cite{Ojala02}.

We write GLC$^k$, GLD$^k$ and BP$^k$ to indicate order $k$ grey level co-occurrence,
grey level difference and binary pattern features, respectively.

\paragraph{Binary equality features}
Complementary to BPs, we considered `binary equality' (BE) feature functions 
which test whether pixels are within an equality threshold $c$ of each other
(dependent on the grey level range)
defined as
\[
  f_\textup{BE}(\Imagepix_{0}, \ldots, \Imagepix_{d}) :=
     \sum_{1\le i \le d} 2^{i-1}\llbracket|\Imagepix_{0} - \Imagepix_{i}| \leq c \rrbracket.
\]
We supposed these might be more suitable for describing
flat regions of an image, where any amount of noise causes $f_\textup{BP}$
to produce evenly distributed random values.
For synthesis purposes, these features must be used in heterogeneous models paired with
other features that break the symmetry between an image and its inverse (GLD features
fail to do this for images with both 180$^\circ$ rotational symmetry and symmetric histograms).
In our experiments we found that if a texture contains flat regions then a number of
BE features would be selected by the nesting procedure, otherwise they were hardly selected.
We found them to have similar descriptive power to BP statistics,
but had additional failure conditions due to the necessary symmetry breaking, so they were not used further.

\subsection{Building up features}
\label{sec:selectors}

As a base model, we used a MGRF with
a first order factor to model marginal statistics
and the two nearest neighbour
GLD features
with offsets $ ((0,0), (1,0))$ and $((0,0), (0,1))$.
As an exception, marginal potentials were not used for the inpainting experiment in Section \ref{sec:inpainting},
which improved the models' ability to match the contrast of the surrounding frame.
However for unguided synthesis matching the original histogram is desirable.

All features were constrained to clique families with at most 40
pixels distance between two points.
The initial candidate set (selector) $C^1$ was always comprised
of all GLD$^2$ features within this circular window.
Three GLD$^2$ features are added at a time to speed learning.
Thereafter different  possibilities for $C^2$ were considered as listed below.

\begin{itemize}
\item
  `Combined' BP$^5$ features built out of the set of characteristic offsets $\{r_i\}$ occurring by the
  previously selected GLD$^2$ and BP$^5$ features. Each possible selection of two offsets $r_1, r_2$
  and for each $i = 1,2$ each choice of either using mirrored offsets $r_i, -r_i$ or halved offsets
  $r_i/2, -r_i/2$ provided the set of four-offset candidates.
  BP$^5$ features were also added two at a time.
\item
  `Conjoined' BP$^9$ features built by selecting (by exhaustive search) four BP$^3$ features
  with symmetric pairs of offsets $(r_i, -r_i)$ of maximal JSD between training and sample histograms
  and then combining them into a single clique family with shape $(r_1, -r_1, \ldots, r_4, -r_4)$.
  Examples can be seen in Figure \ref{fig:cliques1}.
\item
  `Jagged star' (jag-star) BP$^k$ features (for $k$ = 9,13) with 
  $k-1$ surrounding pixels alternately and evenly spaced on
  two circles of radii $d_0$, $d_1$ centred on the central pixel, with offsets at %
  \[
  \begin{bmatrix} x_i \\ y_i \end{bmatrix}
  =
  d_{i\:\textup{mod}\:2}
  \begin{bmatrix}
    \cos(\frac{2\pi i}{k-1} + \phi) \\
    \sin(\frac{2\pi i}{k-1} + \phi)
  \end{bmatrix},
  \]
  where $d_{i\:\textup{mod}\:2}$ is $d_1$ if $i$ is odd and $d_0$ if $i$ is even.
  These configurations include square as well as circular patterns.
  Considering a dense subset of rotations and radii from 1 to 20 pixels provided a fixed set of 2638 candidates.
  Examples can be seen in Figure \ref{fig:cliques1}.
\item
  Linear filters.
  This is a fixed bank of Gabor wavelets and Laplacian of Gaussians identical to those used in FRAME~\cite{FRAME} except that we
  increased the number of Gabor wavelet orientations from 6 to 10, and restricted sizes of the filters
  to prevent excessively long running times.
  We employed Gabor wavelets with wavelengths only up to 6 pixels (filters of size $17\times17$)
  rather than 12.
  In total there were 64 filters.

\end{itemize}

\begin{figure*}[htb]
\centering
  \begin{tabular}{@{}c@{}}
  (a)~\includegraphics[width=0.95\linewidth]{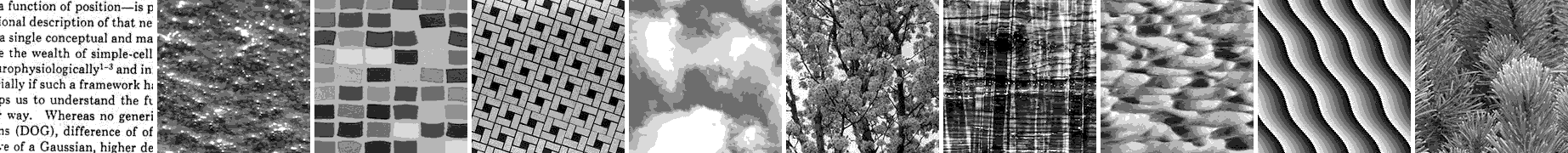} \\
  (b)~\includegraphics[width=0.95\linewidth]{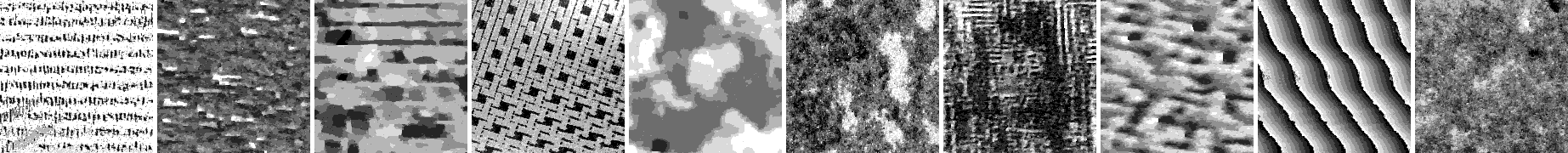} \\
  (c)~\includegraphics[width=0.95\linewidth]{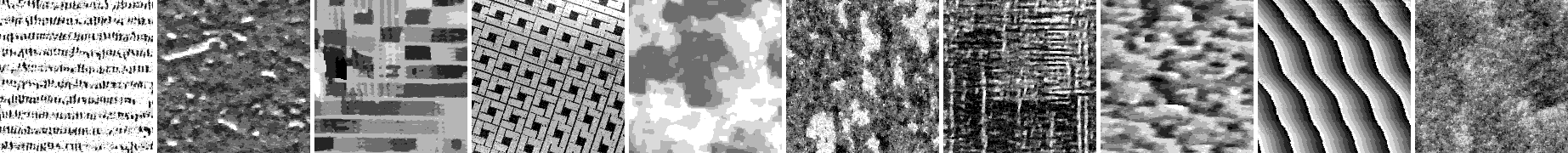} \\
  (d)~\includegraphics[width=0.95\linewidth]{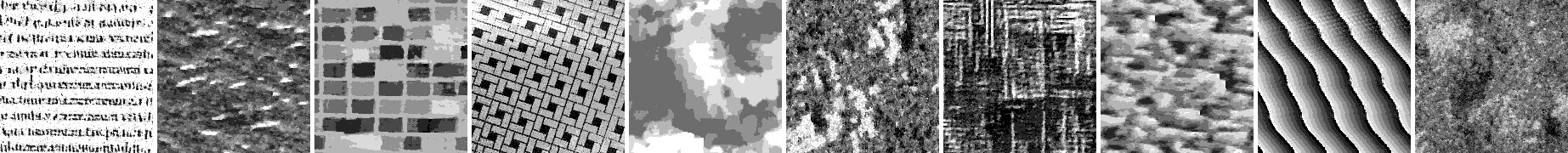} \\
  (e)~\includegraphics[width=0.95\linewidth]{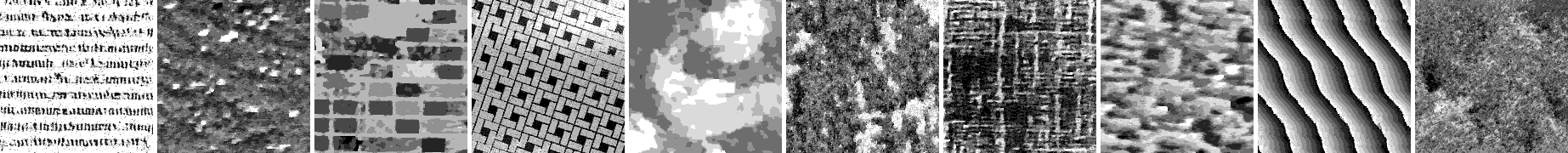} \\
  (f)~\includegraphics[width=0.95\linewidth]{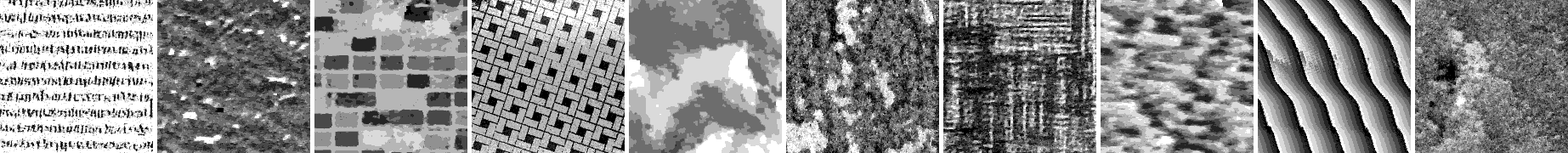} \\
  (g)~\includegraphics[width=0.95\linewidth]{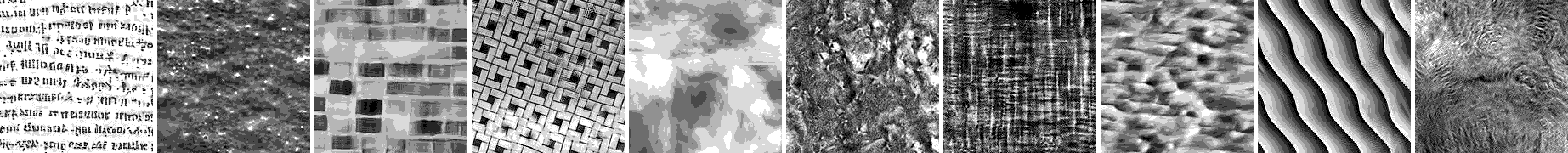} \\
  \end{tabular}

  \caption{
    Synthesis results with models with differing kinds of potentials:
    (a) original textures; 
    (b) second-order GLDs (GLD\textsuperscript{2});
    (c) GLD\textsuperscript{2}+ `combined' BP$^5$; 
    (d) GLD\textsuperscript{2}+`conjoined' BP\textsuperscript{9}; 
    (e) GLD\textsuperscript{2}+jag-star BP\textsuperscript{9};
    (f) GLD\textsuperscript{2}+jag-star BP\textsuperscript{13}; and
    (g) Portilla and Simoncelli's results.
  }
  \label{fig:synth1}
\end{figure*}

\section{Experiments}
\label{sec:experiments}

Generative image models have often been evaluated by measuring performance of
image denoising and inpainting when the model is used as the prior.
But this is an indirect and incomplete method of evaluation.
Instead we mainly compare different classes of models
by inspection of synthesis results.
Experiments were conducted with
a set of grey-scale digitised
photographs of natural and approximately spatially homogeneous textures 
sourced from several popular  databases.
Textures were selected which were
diverse, difficult to model,
homogeneous
and without periodicity or
other features on a scale longer than 40 pixels
(all images were kept at original scales, with the exception of the
inpainting experiment in Section \ref{sec:inpainting}).
The databases used were
the popular Brodatz photoalbum~\cite{Brodatz};
the ``NewTex'' dataset of natural textures
included with
MeasTex~\cite{Meastex-webpage},
a framework for standardised
texture discriminator evaluation; and
the MIT VisTex
database~\cite{VisTex-webpage},
and images collected by the NYU Laboratory for Computational Vision%
~\cite{Portilla00-webpage}.

Textures were subjectively categorised into three classes
according to structure:
stochastic (those apparently described by only simple local interactions, e.g. sand, water), near-regular (regularly tiled but with random defects, e.g. weaves),
and irregular (containing large scale elements with unpredictable placement or shapes, e.g. marbles).

Each image was
quantised to $Q=8$ grey levels.
With some exceptions (e.g. when comparing to other results)
the images were quantised
using contrast-limited adaptive histogram equalization
\cite{Zuiderveld94}
with $16 \times 16$ tiles and a contrast clipping limit of $0.03$.
Adaptive histogram equalization
mostly avoids the situation where most of the image
is mapped to only one or two
grey levels, and also lessens shadows and gradients,
which hinder the recovery of long range interactions.
The low $Q$ value used speeds up Gibbs sampling.

Source code for all experiments
is freely available from the website
accompanying the paper, together with supplementary
material including results for a large number of additional textures%
\footnote{
  \url{http://www.ivs.auckland.ac.nz/texture_modelling}.
}.

\begin{figure*}[htbp!]
\centering
  \begin{tabular}{@{}c@{}}
  (a)~\includegraphics[width=0.85\linewidth]{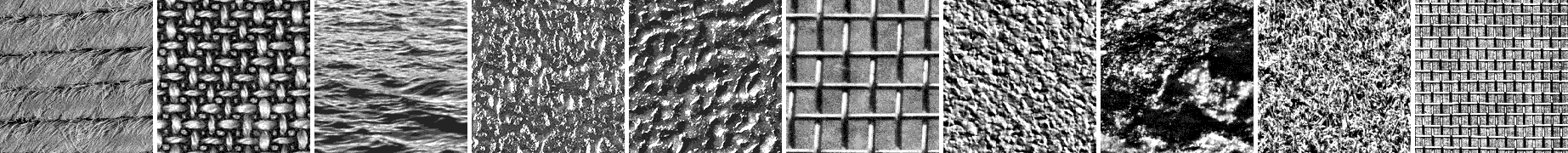} \\
  (b)~\includegraphics[width=0.85\linewidth]{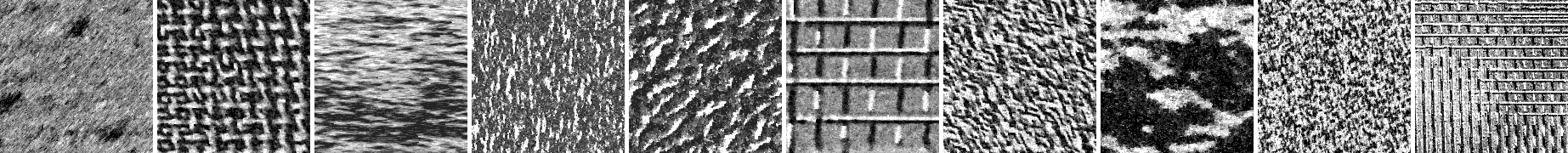} \\
  (c)~\includegraphics[width=0.85\linewidth]{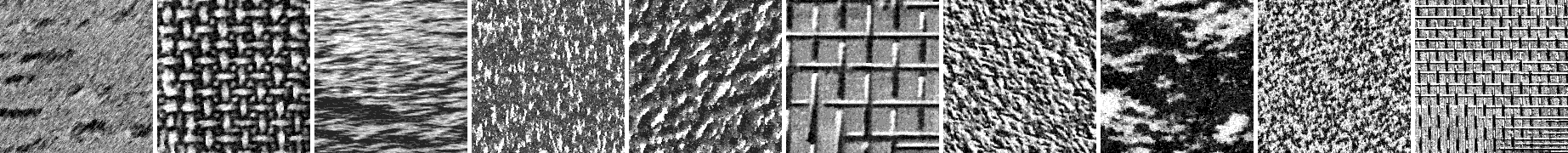} \\
  (d)~\includegraphics[width=0.85\linewidth]{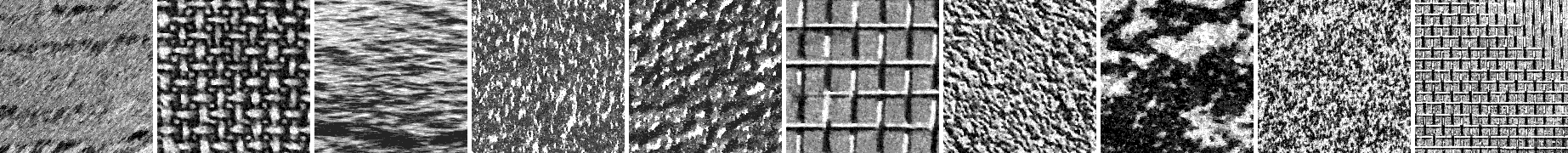} \\
  (e)~\includegraphics[width=0.85\linewidth]{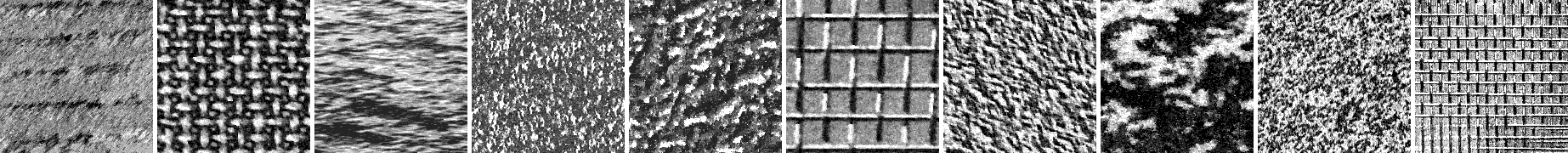} \\
  (f)~\includegraphics[width=0.85\linewidth]{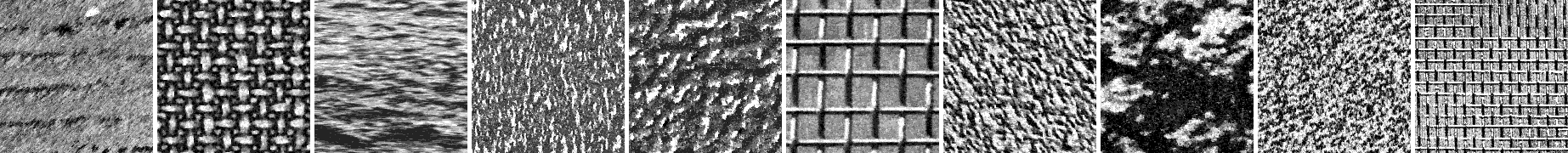} \\
  (g)~\includegraphics[width=0.85\linewidth]{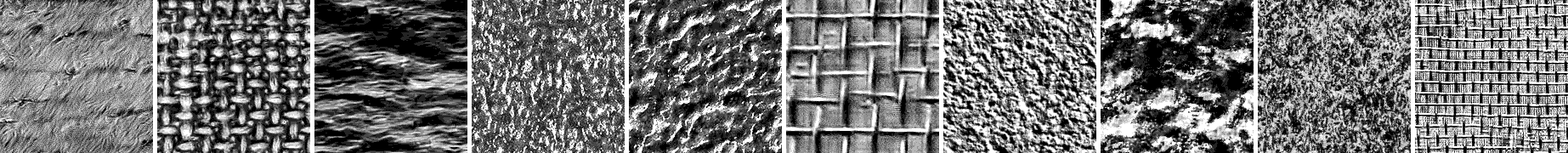} \\
  (a)~\includegraphics[width=0.85\linewidth]{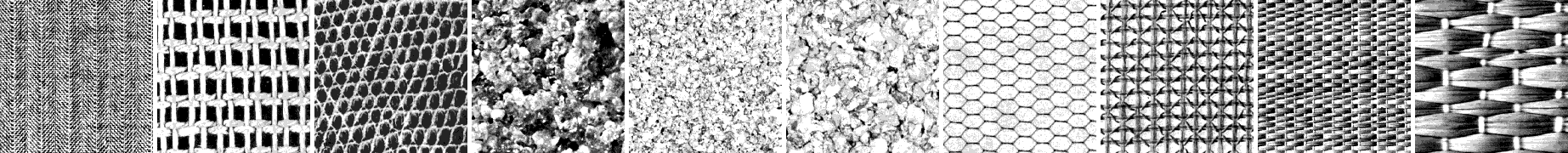} \\
  (b)~\includegraphics[width=0.85\linewidth]{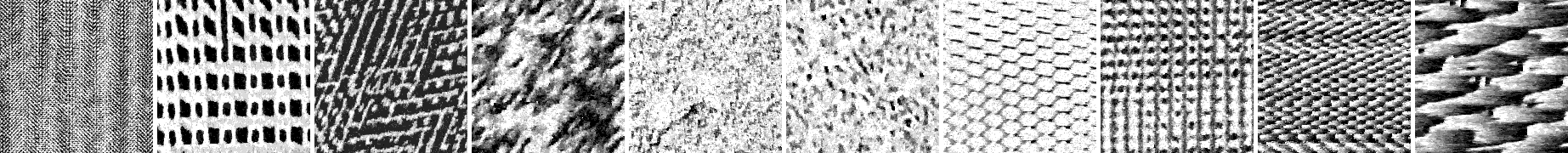} \\
  (c)~\includegraphics[width=0.85\linewidth]{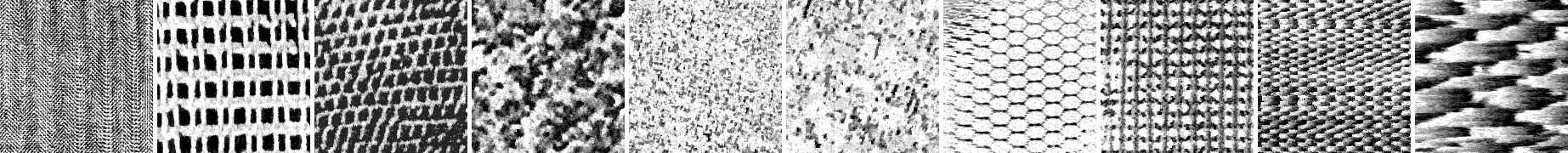} \\
  (d)~\includegraphics[width=0.85\linewidth]{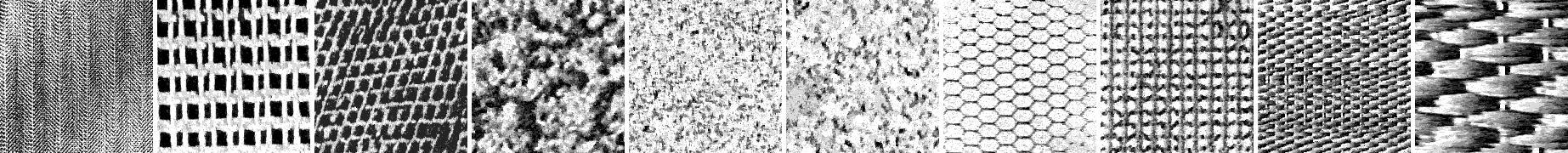} \\
  (e)~\includegraphics[width=0.85\linewidth]{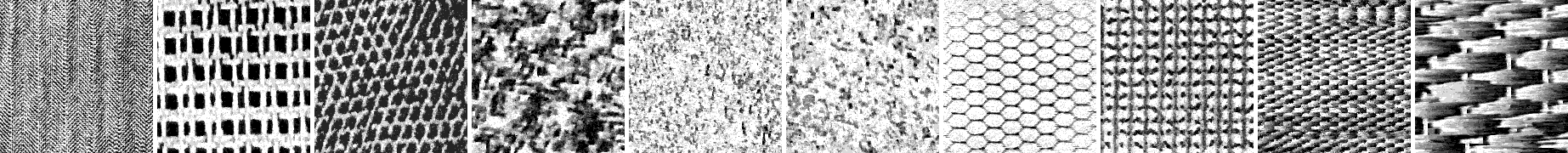} \\
  (f)~\includegraphics[width=0.85\linewidth]{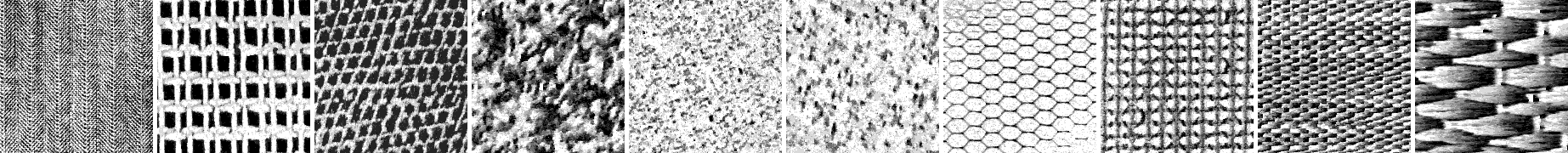} \\
  (g)~\includegraphics[width=0.85\linewidth]{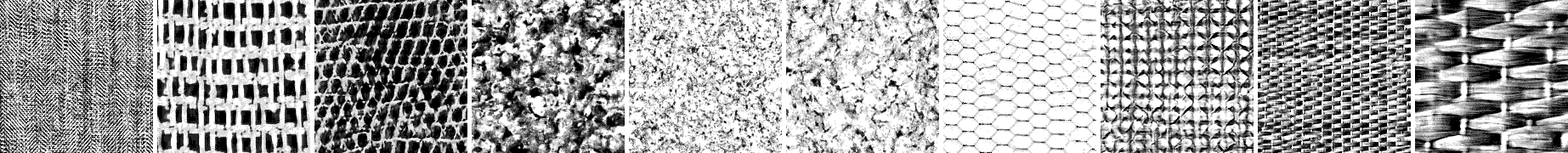} \\
  \end{tabular}
  \caption{
    Synthesis results continued: see Figure \ref{fig:synth1}.
    }
  \label{fig:synth3}
\end{figure*}

\begin{figure*}[htb]
\ifdefined\includeallfigs
  \centerline{
    \includegraphics[width=0.95\linewidth]{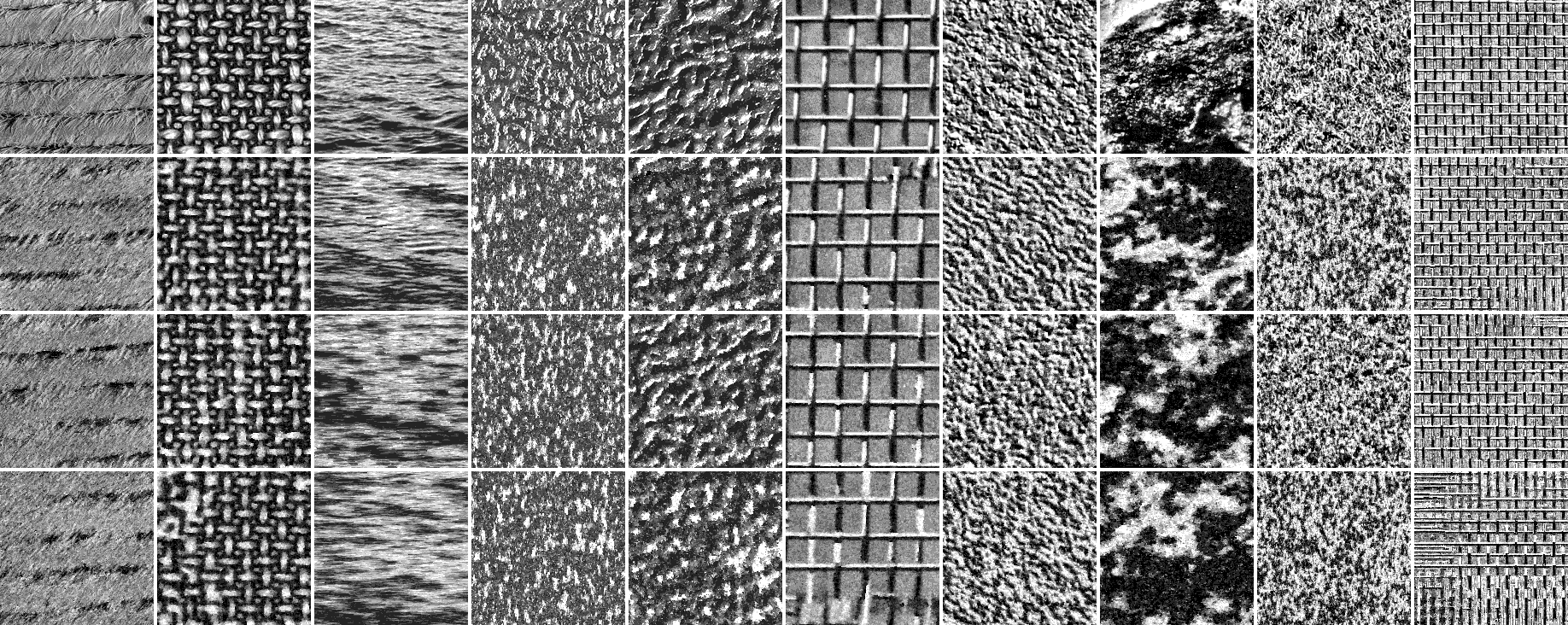}
  }
\fi

  \caption{
    Variants on jag-star BP models. Rows are:
    \emph{First}: training images.
    \emph{Second}: GLD$^2$ + jag-star BP$^9$ models. %
    \emph{Third}: GLC$^2$ + jag-star BP$^9$ models.
    \emph{Fourth}: jag-star BP$^9$ models without any pairwise potentials (except for initial adjacent ones).
  }
  \label{fig:pairwise-comp}
\end{figure*}

\begin{figure}[htb]
\centering
\includegraphics[width=0.9\columnwidth]{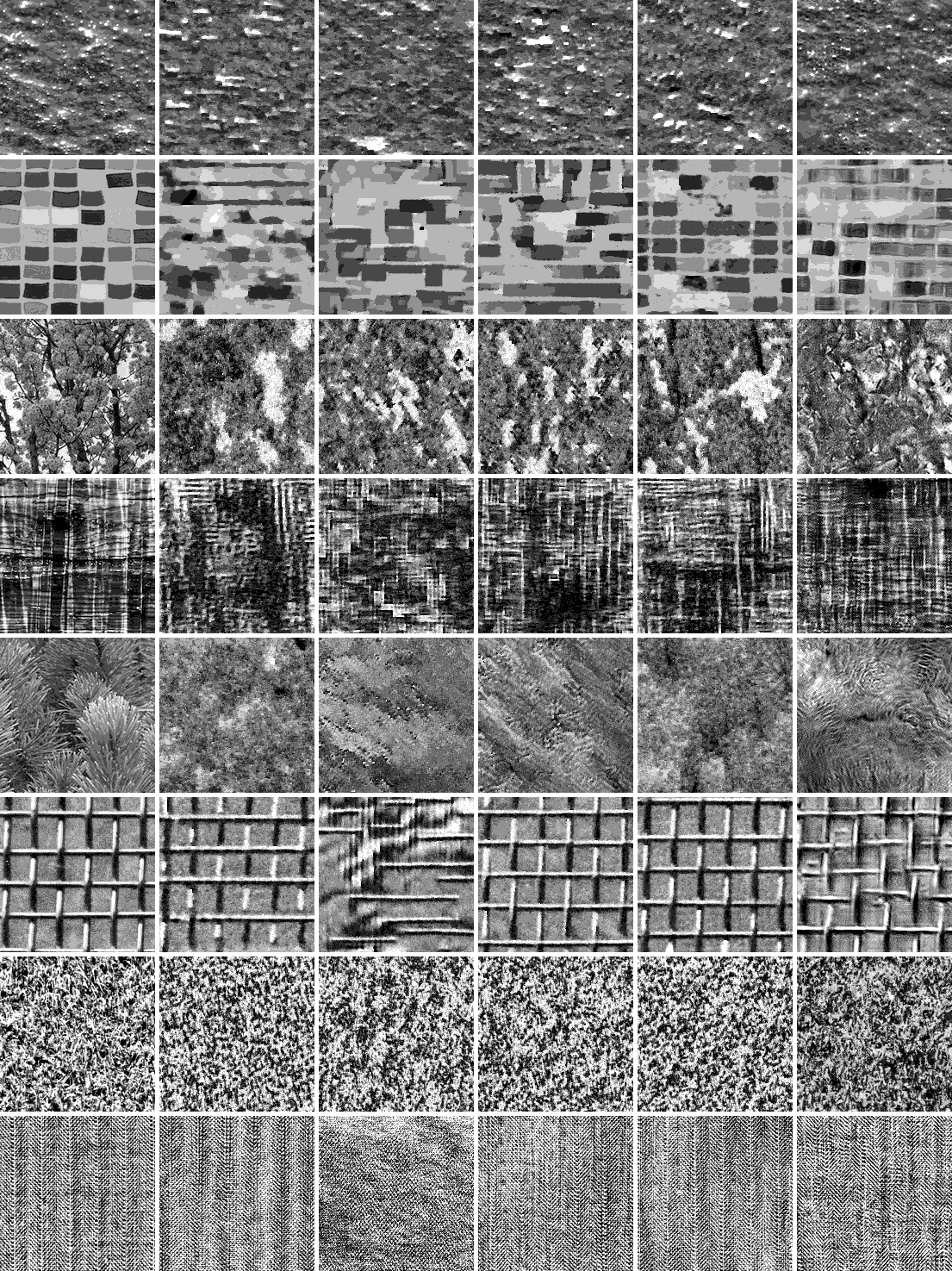}
\caption{
Comparison of synthesis results using FRAME linear filters
to other models. Column 1: original textures; columns 2 to 5: models with GLD$^2$, linear filters, GLD$^2$ + linear filters, and GLD$^2$ + jag-star BP$^{13}$ features, respectively.
}
\label{fig:frame}
\end{figure}

\subsection{Synthesis}
\label{sec:synth}

As a baseline we compare synthesis results to the texture
analysis and synthesis algorithm by Portilla and Simoncelli \cite{Portilla00}
(downloadable from \cite{Portilla00-webpage}),
which in addition to being freely available, is today 
still one of the most successful attempts to
model texture statistically, and has recently been extended
to colour textures.
Their approach uses iterated projections to attempt
to produce images matching certain first- and second-order statistics of wavelet
responses.
On the other hand,
all of the recent filter-based
MRF texture models \cite{Heess09,Ranzato10,Kivinen12,Luo13,Gao14}
have only been demonstrated on a few simple highly regular textures with
short repetition lengths and texton scales.
Unfortunately the lack of difficult synthesis examples
or source code
for these previously published approaches hinders
comparison,
although \cite{Luo13} achieved visually better results
than us on the
difficult D103/D104 textures (see Figure \ref{fig:synth-comparison}) and D76
by using three layers of hundreds of filters in a deep belief network.
In theory the purpose of these layers was to stabilise learning
by using tiled-convolutional rather than fully-convolutional
filters, which for simpler models is not necessary \cite{Gao14}.

Figures \ref{fig:synth1} and \ref{fig:synth3}
show $180\times180$ synthesis results
for 30 of the most diverse and interesting examples
modelled using each of the alternative classes of BP features described in Section \ref{sec:selectors}.
Training images were $256\times 256$ in size.

Synthesis results for the different models under comparison
were evaluated by a panel of observers.
For each of 20 textures (a subset of the 30 shown)
six synthesised images from different classes of models was presented to each person along with
the original training image, and they were tasked with placing the
synthesis results in order from best to worst%
\footnote{
  This questionnaire is still available online at \url{http://www.ivs.auckland.ac.nz/texture_modelling/}.
}.
The observers were
instructed to make rapid decisions rather
than perform careful inspection.

Table \ref{tab:synth-results-human-summary} presents
the average rankings of each of the model types
against the other five options, summarised by texture category.
Additionally Table \ref{tab:synth-results-human-rank1} reports how frequently
each class of model gave the best-ranked synthesis result.
The BP$^5$ features outperforming the 9th order ones on
stochastic textures may be an artifact of the earlier stopping rule,
which usually excluded the higher order ones altogether on these textures.

From the figures and tables, it can be seen that the MGRFs perform
well on near-regular textures, as tesselation is easily handled,
and on stochastic textures, for which the apparent dependency of the
actual texture is on a short scale and thus a natural fit for a MGRF.
However, it is not easy to separate the performance of the different classes of models.
On stochastic textures the method of Portilla and Simoncelli 
usually but not always dominates,
apparently most significantly because it captures
high frequency aspects and details (such as thin branches and speckles) that ours do not.
It is often
able to produce nice results for irregular textures too,
however, it often violates fixed tesselation offsets
and fails to represent complex textons.
On those cases the simple and cheap GLD potentials can be very effective.
In addition, Gibbs (or CSA) sampling from a MGRF adds unwanted high frequency noise.
Complex irregular textures such as tiles in column 3 of Figure~\ref{fig:synth1}
are the most difficult for all methods,
and in these cases hidden variables to distinguish different local
contexts, as in gated MRFs, are probably necessary to visually
replicate the texture.

In the tables the `combined' 5\textsuperscript{th} order BPs
are seen to do very well on irregular textures, despite being low order
and only having 16 parameters per feature.
This provides promising evidence that piecing together high order features
out of low order ones is a workable technique.
However a number of other schemes for building up to higher orders that we
attempted, not documented here, provided bad results.
This motivated the investigation of the jag-star features as simpler and
less brittle alternatives.
Previous independent experiments \cite{Versteegen14} also favoured the `combined' features, providing more evidence for a real effect.
No signficant difference between the 9\textsuperscript{th} and 13\textsuperscript{th}
order jag-star features is visible,
which is provides some evidence that the large number of parameter in the latter models --- 4096 parameters
per BP feature --- is not unworkable.
Results for star BP's (not shown) were close to those
for jag-star, but generally inferior.

\begin{table}[!tb]
  \caption{
    Averages and standard deviations of human rankings of texture synthesis results by texture category:
    1 was best and 6 worst.
    All models except Portilla \& Simoncelli \cite{Portilla00} include GLD potentials.
  }
  \label{tab:synth-results-human-summary}
  \centering

  \setlength{\tabcolsep}{5pt}
  \begin{tabular}{lccccc}
    \toprule
  &  Stochastic  &  Near-regular  &  Irregular  &  Total \\

  \midrule

Lin. filters  &  3.0$\pm$1.3  &  4.1$\pm$1.2  &  4.0$\pm$1.4  &  3.8$\pm$1.3 \\

Comb. BP$^5$  &  4.1$\pm$1.2  &  2.0$\pm$1.0  &  3.6$\pm$1.3  &  3.2$\pm$1.2 \\

Conj. BP$^9$   &  4.4$\pm$1.3  &  3.4$\pm$1.2  &  3.0$\pm$1.2  &  3.5$\pm$1.2 \\

JStar BP$^{9}$    &  4.0$\pm$1.5  &  3.8$\pm$1.2  &  3.7$\pm$1.2  &  3.8$\pm$1.3 \\

JStar BP$^{13}$    &  4.0$\pm$1.3  &  3.3$\pm$1.0  &  3.1$\pm$1.1  &  3.4$\pm$1.1 \\

P\&{}S \cite{Portilla00}  &  1.5$\pm$1.0  &  4.3$\pm$1.3  &  3.7$\pm$1.4  &  3.3$\pm$1.3 \\

\bottomrule
  \end{tabular}
\end{table}

\begin{table}[!tb]
  \caption{
    Fraction of the time that an image synthesised by each class of texture model
    was ranked best by a judge out of the 6 possibilities,
    split by texture category.
    All models except Portilla \& Simoncelli \cite{Portilla00} include GLD potentials.
  }
  \label{tab:synth-results-human-rank1}
  \centering

  \setlength{\tabcolsep}{6pt}
  \begin{tabular}{lccccc}
    \toprule
  &  Stochastic  &  Near-regular  &  Irregular  &  Total \\

  \midrule

Lin. filters  &  6\%  &  9\%  &  11\%  &  9\% \\

Comb. BP$^5$  &  2\%  &  47\%  &  11\%  &  21\% \\

Conj. BP$^9$   &  0\%  &  12\%  &  22\%  &  13\% \\

JStar BP$^{9}$    &  6\%  &  7\%  &  5\%  &  6\% \\

JStar BP$^{13}$    &  2\%  &  12\%  &  26\%  &  15\% \\

P\&{}S \cite{Portilla00}  &  82\%  &  9\%  &  23\%  &  33\% \\
\bottomrule
  \end{tabular}
\end{table}

\setlength\columnsep{10pt}

\begin{figure}[htb]

  \makebox[\columnwidth][c]{
  \includegraphics[width=\columnwidth]{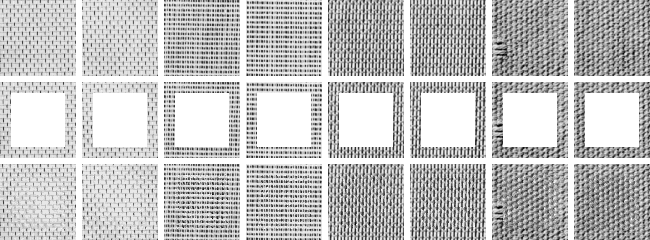}
  }
  \makebox[\columnwidth][c]{
    \includegraphics[width=\columnwidth]{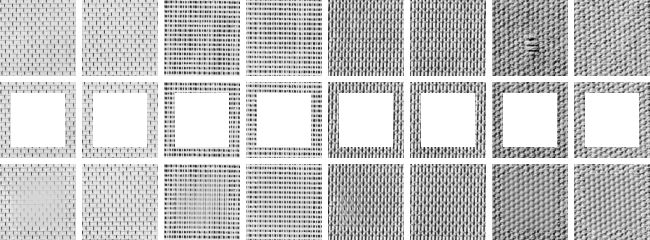}
  }
  \caption{  \label{fig:inpainting}
    Inpainting examples for a 9th-order jag-star model of D6, D21, D53 and D77.
    The left hand column of each pair shows the worst inpainting result over 20 repetitions,
    the right hand column the best result.
    The top block of results is for simple inpainting using Gibbs sampling,
    the bottom block shows a smoothed result by averaging 50 consecutive Gibbs iterations.
    The smoothed images have much higher MSSIM scores.
  }
\end{figure}

\subsubsection{GLC vs. GLD features}

We also compared the effect of replacing pairwise GLD features with GLC
features or removing them entirely, as shown in Figure \ref{fig:pairwise-comp}.
Generally the result of excluding pairwise potentials is not very great
(although it is very large when linear filters are used instead, Section \ref{sec:linfilters}),
or even undetectable, as the higher order features can seemingly usually capture
the same statistics, which is against expectations.
Naturally long range GLD features are very helpful for regular textures.
The difference between GLD and GLC features is even less, though
over the whole texture set use of GLC models as a base seem to be slightly more powerful.

\subsubsection{Linear vs. nonlinear filters}
\label{sec:linfilters}
Using BPs for texture modelling was compared with more traditional linear filters, too.
We consider both models
which are close to those used in \cite{FRAME}
(with both  nesting procedure and filter bank being similar),
and models also including GLD$^2$ features.
In either case the initial base model was the same as used in all our other models.

We can see from Figure~\ref{fig:frame}
that small linear filters alone of course can not reconstruct
textures whenever textural elements are larger than the filter wavelength,
although larger objects can still be partially captured, such as the tiles
in row 2, column 3.
Augmenting the filters with long range GLD potentials prevents this failure
case, while the linear filters perform much better than pairwise features alone.
Unexpectedly with the addition of these long range potentials
the synthesised images do not normally appear distinct from those composed of
BP features, with very similar results across all texture types.
For example the speckles in the first row texture are disappointingly not
captured, indicating a deficiency in the fixed filterbank.
However in row 5 (pine trees)
we do see some ripples in the images clearly captured by wavelets,
including Portilla \& Simoncelli's analysis, but not by BP features.
Additionally from Table \ref{tab:synth-results-human-summary}
it appears that the FRAME linear filters outperform BPs on stochastic
textures, as might be expected, although far weaker than the comprehensive set of
wavelet statistics used in \cite{Portilla00};
but on the other hand are outperformed by BPs on textures with more structure.

Despite restricting the filter sizes to $17\times17$,
model learning times with the filters were still 30 minutes on average
(our implementation in C++ is single-threaded), several times that of the BP models.

\begin{figure*}[htbp]
\centering
  \begin{tabular}{@{}l@{~}c@{~}c@{~}c@{~}c@{~}c@{~}c@{~}c@{~}c@{}}
    & D6 & D21 & D53 & D77 & D4 & D14 & D68 & D103 \\
(a) %
& \includegraphics[width=0.106\linewidth]{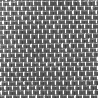}
& \includegraphics[width=0.106\linewidth]{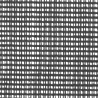}
& \includegraphics[width=0.106\linewidth]{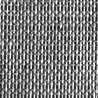}
& \includegraphics[width=0.106\linewidth]{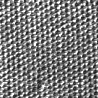}
& \includegraphics[width=0.106\linewidth]{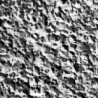}
& \includegraphics[width=0.106\linewidth]{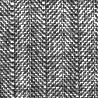}
& \includegraphics[width=0.106\linewidth]{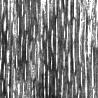}
& \includegraphics[width=0.106\linewidth]{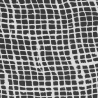}\\
(b) %
& \includegraphics[width=0.106\linewidth]{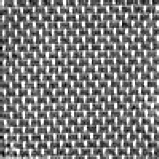}
& \includegraphics[width=0.106\linewidth]{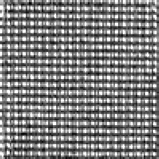}
& \includegraphics[width=0.106\linewidth]{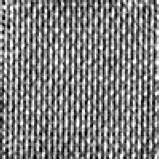}
& \includegraphics[width=0.106\linewidth]{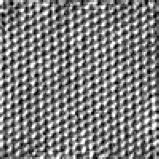}
& \includegraphics[width=0.106\linewidth]{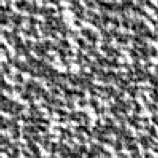}
& \includegraphics[width=0.106\linewidth]{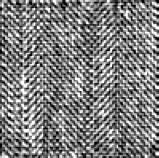}
& \includegraphics[width=0.106\linewidth]{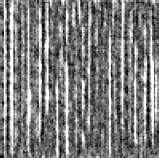}
& \includegraphics[width=0.106\linewidth]{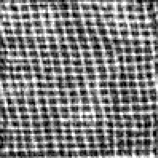}\\
(c) %
& \includegraphics[width=0.106\linewidth]{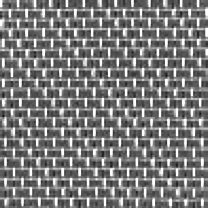}
& \includegraphics[width=0.106\linewidth]{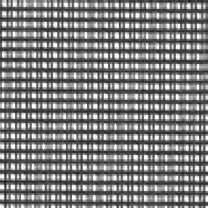}
& \includegraphics[width=0.106\linewidth]{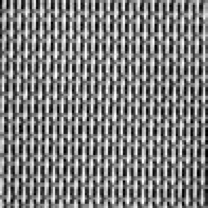}
& \includegraphics[width=0.106\linewidth]{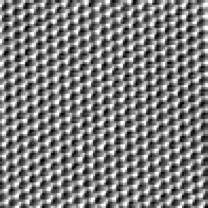}
& \includegraphics[width=0.106\linewidth]{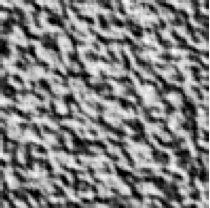}
& \includegraphics[width=0.106\linewidth]{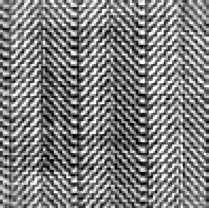}
& \includegraphics[width=0.106\linewidth]{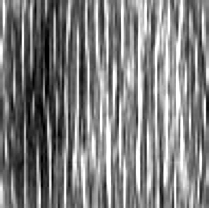}
& \includegraphics[width=0.106\linewidth]{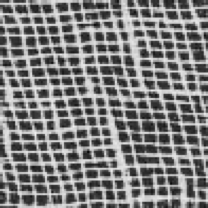}\\
\midrule
(d) %
& \includegraphics[width=0.106\linewidth]{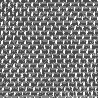}
& \includegraphics[width=0.106\linewidth]{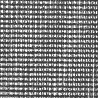}
& \includegraphics[width=0.106\linewidth]{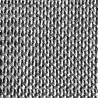}
& \includegraphics[width=0.106\linewidth]{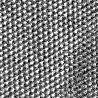}
& \includegraphics[width=0.106\linewidth]{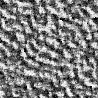}
& \includegraphics[width=0.106\linewidth]{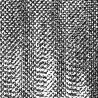}
& \includegraphics[width=0.106\linewidth]{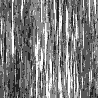}
& \includegraphics[width=0.106\linewidth]{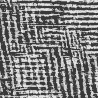}\\
(e) %
& \includegraphics[width=0.106\linewidth]{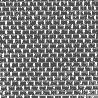}
& \includegraphics[width=0.106\linewidth]{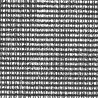}
& \includegraphics[width=0.106\linewidth]{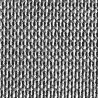}
& \includegraphics[width=0.106\linewidth]{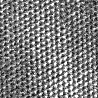}
& \includegraphics[width=0.106\linewidth]{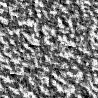}
& \includegraphics[width=0.106\linewidth]{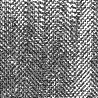}
& \includegraphics[width=0.106\linewidth]{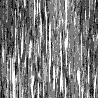}
& \includegraphics[width=0.106\linewidth]{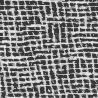}\\
(f) %
& \includegraphics[width=0.106\linewidth]{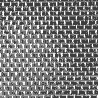}
& \includegraphics[width=0.106\linewidth]{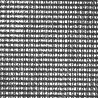}
& \includegraphics[width=0.106\linewidth]{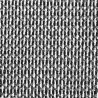}
& \includegraphics[width=0.106\linewidth]{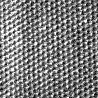}
& \includegraphics[width=0.106\linewidth]{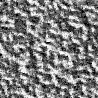}
& \includegraphics[width=0.106\linewidth]{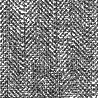}
& \includegraphics[width=0.106\linewidth]{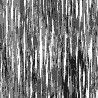}
& \includegraphics[width=0.106\linewidth]{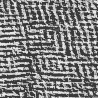}\\
(g) %
& \includegraphics[width=0.106\linewidth]{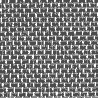}
& \includegraphics[width=0.106\linewidth]{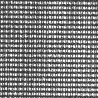}
& \includegraphics[width=0.106\linewidth]{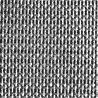}
& \includegraphics[width=0.106\linewidth]{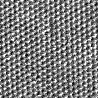}
& \includegraphics[width=0.106\linewidth]{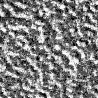}
& \includegraphics[width=0.106\linewidth]{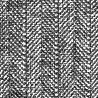}
& \includegraphics[width=0.106\linewidth]{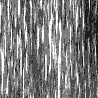}
& \includegraphics[width=0.106\linewidth]{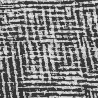}\\
  \end{tabular}
  \caption{\label{fig:synth-comparison}
    Comparison of our synthesis results against previously published works:
    (a)the eight original $98\times 98$ Brodatz textures; (b) results of the Multi-Tm
    algorithm~\cite{Kivinen12}; (c) results of the 2-layer DBM~\cite{Luo13}; (d) -- (g) our
    results with GLD\textsuperscript{2}, GLD\textsuperscript{2} + combined BP\textsuperscript{5}, 
    GLD\textsuperscript{2} + conjoin BP\textsuperscript{9}, and GLD\textsuperscript{2}
    + jagstar BP\textsuperscript{13}. The original images are 
     scaled and with grey levels reversed as in the previous works.
    Each image has been individually renormalised in order to erase contrast
    differences in the original textures.}
\end{figure*}

\subsection{Inpainting}
\label{sec:inpainting}

\begin{table*}[htb]
  \centering
  \setlength{\tabcolsep}{6pt}

  \begin{tabular}{lcccc}
\toprule
                                           & D6 & D21 & D53 & D77 \\
\midrule
Efros \&{} Leung  \cite{Efros99} & 0.85 $\pm$ 0.03 & 0.86 $\pm$ 0.03 & 0.86 $\pm$ 0.06 & 0.60 $\pm$ 0.08 \\
TmPoT \cite{Kivinen12}                     & 0.86 $\pm$ 0.02 & 0.87 $\pm$ 0.01 & 0.86 $\pm$ 0.02 & 0.77 $\pm$ 0.03 \\
TssRBM \cite{Luo13}                        & 0.89 $\pm$ 0.02 & 0.91 $\pm$ 0.01 & 0.92 $\pm$ 0.02 & 0.76 $\pm$ 0.03 \\
2-layer DBN \cite{Luo13}                   & 0.89 $\pm$ 0.03 & 0.91 $\pm$ 0.02 & 0.92 $\pm$ 0.03 & 0.77 $\pm$ 0.02 \\
cGRBMs \cite{Gao14}                        & 0.91 $\pm$ 0.02 & 0.93 $\pm$ 0.01 & 0.93 $\pm$ 0.01 & 0.78 $\pm$ 0.03 \\

 \midrule

             GLD$^2$ & $0.66 \pm 0.15$ & $0.90 \pm 0.04$ & $0.86 \pm 0.08$ & $0.78 \pm 0.03$ \\
             Linear filters & $0.77 \pm 0.02$ & $0.49 \pm 0.13$ & $0.89 \pm 0.04$ & $0.58 \pm 0.06$ \\
      Combined  BP$^5$ & $0.80 \pm 0.08$ & $0.87 \pm 0.06$ & $0.92 \pm 0.03$ & $0.81 \pm 0.02$ \\
      Conjoined BP$^9$ & $0.77 \pm 0.10$ & $0.86 \pm 0.06$ & $0.90 \pm 0.03$ & $0.80 \pm 0.03$ \\
     Jag-star BP$^9$ & $0.84 \pm 0.08$ & $0.86 \pm 0.07$ & $0.91 \pm 0.03$ & $0.80 \pm 0.03$ \\
  Jag-star BP$^{13}$ & $0.73 \pm 0.11$ & $0.90 \pm 0.05$ & $0.91 \pm 0.03$ & $0.79 \pm 0.03$ \\

\bottomrule
  \end{tabular}
  \caption{
    Results of inpainting evaluation; MSSIM scores against ground truth (\emph{mean $\pm$ standard deviation}).
    All our models (in the bottom half) include pairwise GLD potentials in addition to the higher order ones.
    Results for the Efros \&{} Leung algorithm are taken from \cite{Kivinen12} and used $19 \times 19$ patch sizes.
  }
  \label{tab:inpainting-results}
\end{table*}

Additionally, we compared against results from three recent
leading hierarchical MGRF texture models \cite{Kivinen12,Luo13,Gao14}
which attempted to quantitatively evaluate MGRF texture
models through a  texture inpainting task.
(Heess et al.~\cite{Heess09} used a slightly different variant, so most of their
results are incomparable.)
By focusing on four highly regular textures with small textural feature sizes,
it is reasonable to compare the synthesized pixels in the inpainted
region to the original ones
since they should be nearly aligned.
This is not applicable to other texture types,
so is a very narrow test.

In the task a $76 \times 76$ pixel
piece of Brodatz textures D6, D21, D53, or D77 is provided as an inpainting frame
with a $54 \times 54$ pixel hole cut out of the centre.
An algorithm fills
the hole in a way consistent with the frame,
and the difference against the ground truth is measured
using the
mean structural similarity index (MSSIM) \cite{Wang04}.
MSSIM has been found experimentally to be a good measure of perceptual distance between images.
D6 and D53 were bilinearly scaled to 50\% and D21 and D77 to 75\%;
this allowed the use of small filters by the earlier papers.
The top half of each image was used for training, and the bottom half for testing.
Since the learning procedure is stochastic, the inpainting is repeated (200 times in our case) with random inpainting frames
from test region and the average MSSIM score reported.

Examples of the inpainted frames are shown in Figure \ref{fig:inpainting},
and examples of texture \emph{synthesis} using the models used for inpainting
is shown in Figure~\ref{fig:synth-comparison}, comparing
to previously published synthesis results.
Initially, the MSSIM scores across all models we learnt were very poor,
but Figure~\ref{fig:synth-comparison}
reveals the reason for the bad quantitative results.
All the best previously published models produce very smooth synthesis
and also inpainting (not shown) results,
as if maximising the per-pixel posterior marginals
rather than providing a sample from the texture distribution.
This is especially true for the published works which achieve the
highest MSSIM scores \cite{Gao14}.
On the other hand our models produce typically more realistic-appearing
inpainting and synthesis results, with random deviations.

By averaging over 50 consecutive Gibbs MCMC samples
we now achieve similar scores to previously published algorithms, although visual quality is greatly
decreased by doing so. This is demonstrated in Figure \ref{fig:inpainting}.
The MSSIM scores are presented in Table \ref{tab:inpainting-results},
alongside best previously published results
and the Efros and Leung texture synthesis algorithm \cite{Efros99} as a baseline.
For this experiment we used $Q=16$ rather than $Q=8$ grey levels.
The comparison of ground truth with 256 grey levels against
models using only 16 levels gives a slight penalty against the previous works.
For some unknown reason all models attempted frequently failed to properly
inpaint D6, resulting in low scores for this texture.
The full resolution version of D6 is also unusually difficult to synthesize
for all models.

\section{Conclusion}

While Markov random fields have long been applied to
image modelling, there has been little prior investigation
in this context
into practical learning of interaction structure,
with until recently \cite{Liu14:1} apparently only 
2nd to 4th order MGRFs previously learned.
The particular model nesting approach applied in this paper
has the same theoretical foundation on maximum entropy distributions
as previous sequential structure learning approaches,
but emphasizes approximation and speed by using CSA.
Aside from this, the strength of model nesting is in creating heterogeneous models,
and allowing composition of feature functions.
Mixing dissimilar potentials in a MGRF typically shows immediate improvements.

Furthermore, for nearly two decades MGRF texture and image
models based on square linear filters have been almost exclusively
studied due to their superior performance over the primitive pairwise
models that preceded them.
However, we have shown that other types of high-order
features, such as the generic binary pattern features introduced in this paper, %
are viable alternatives to linear filters.
These BP features have several advantages over filters.
They are offset and nearly contrast invariant, %
and are still powerful even when of low order.
The presented synthesis and inpainting experiments showed that even BP feature of only 5\textsuperscript{th}
order are sufficient to capture many important visual aspects of textures
and are capable of outperforming the previously best approaches,
thanks to building features up out of characteristic
low-order cliques.
These relatively low orders reduce computation costs,
while no filter coefficients need to be selected or learned.

The purpose of our synthesis experiments is not
to challenge the mainstream highly effective and efficient
texture synthesis algorithms but to validate our probabilistic models.
There are however still many textures which
our models have failed to  capture,
especially complex irregular ones, as well as losing
fine details in general.
In order to tackle these in future we plan to extend our
models by including
additional layers of feature selectors for diverse types of
statistics, particularly
co-occurrences between the outputs of feature
functions at previous levels, as has been used in several texture models
which used filters as features.
A stopping rule for the nesting procedure which is robust to the highly
variable quality of CSA samples would also be important if more feature sets are used,
to keep the number of nesting iterations low.
In future work we also hope to improve the generalisation ability of
the models, such as with invariance to small deformations of the image.
This is closely related to the selection of complex composite features
which can describe more powerful abstractions.

\section*{\refname}
\bibliography{references,urls}{}
\bibliographystyle{elsarticle-num}

\end{document}